\documentclass{article} % For LaTeX2e
\usepackage{enumitem}
\usepackage{needspace}
\usepackage{subcaption} 
\usepackage{listings}
\PassOptionsToPackage{table}{xcolor}
\usepackage{siunitx}
\usepackage{colortbl}
\usepackage{textcomp}
\usepackage{tikz}
\usepackage[table]{xcolor}
\definecolor{lightblue}{RGB}{220, 230, 241}
\usepackage{tabularx}
\newcolumntype{Y}{>{\centering\arraybackslash}X}  % centered X‐column
\definecolor{codegray}{rgb}{0.5,0.5,0.5}
\definecolor{codepurple}{rgb}{0.58,0,0.82}
\definecolor{backcolour}{rgb}{0.95,0.95,0.95}

\lstdefinestyle{python}{
    backgroundcolor=\color{backcolour},
    commentstyle=\color{codegray}\ttfamily,
    keywordstyle=\color{blue}\bfseries,
    numberstyle=\tiny\color{gray},
    stringstyle=\color{codepurple},
    basicstyle=\ttfamily\footnotesize,
    breakatwhitespace=false,
    breaklines=true,
    captionpos=b,
    keepspaces=true,
    numbers=left,
    numbersep=5pt,
    showspaces=false,
    showstringspaces=false,
    showtabs=false,
    tabsize=4,
    language=Python
}

\usetikzlibrary{arrows.meta, positioning, shapes.multipart, calc}
\tikzset{
  block/.style={
    draw,
    thick,
    rounded corners=2pt,
    minimum width=2.5cm,
    minimum height=1cm,
    font=\small\sffamily
  },
  predictor/.style={block, fill=orange!40},
  encoder/.style={block, fill=blue!20},
  module/.style={block, fill=red!30},
  source/.style={block, fill=gray!10},
  arrow/.style={-Latex, thick},
}

\usepackage{graphicx}
\usepackage{bbm}
\usepackage[colorinlistoftodos]{todonotes}
\usepackage[inkscapelatex=false]{svg}
\usepackage[bottom]{footmisc}
\usepackage[font=small,labelfont=bf]{caption}
\usepackage{tablefootnote}
\usepackage{textcomp}
\usepackage{mathtools}
\usepackage{amsmath}     % For mathematical expressions
\usepackage{amssymb}     % For symbols like \mathbb{R}
\usepackage{algorithm}   % Defines the 'algorithm' float environment
\usepackage{algpseudocode}
\usepackage[preprint]{neurips_2025}

\usepackage[utf8]{inputenc} % allow utf-8 input
\usepackage[T1]{fontenc}    % use 8-bit T1 fonts

\usepackage{amsmath,amsfonts,bm}

\def\eqref#1{equation~\ref{#1}}
\def\1{\bm{1}}

\DeclareMathAlphabet{\mathsfit}{\encodingdefault}{\sfdefault}{m}{sl}
\SetMathAlphabet{\mathsfit}{bold}{\encodingdefault}{\sfdefault}{bx}{n}

\usepackage{multirow}
\usepackage{pifont}

\newcommand{\cmark}{\ding{51}} % ✓
\newcommand{\xmark}{\ding{55}} % ✗

\usepackage{caption} 
\usepackage{hyperref}
\usepackage{url}
\usepackage{float}
\usepackage{booktabs}
\usepackage{amssymb}
\usepackage{amsmath}
\usepackage{tikz}
\usepackage{microtype}      % microtypography
\usepackage{amsfonts}       % blackboard math symbols
\usepackage[font=footnotesize,labelformat=simple]{subcaption}

\usepackage{nicefrac}       % compact symbols for 1/2, etc.
\usepackage{cleveref}

\usetikzlibrary{shapes.geometric, arrows}

\title{Time to Embed: Unlocking Foundation Models for Time Series with Channel Descriptions}

\author{%
  Utsav Dutta \\
  C3 AI \\
  1400 Seaport Blvd\\ 
  Redwood City, CA 94063\\
  \texttt{utsav.dutta@c3.ai} \\
  \And
  Sina Khoshfetrat Pakazad \\
  C3 AI \\
  1400 Seaport Blvd\\ 
  Redwood City, CA 94063\\
  \texttt{sina.pakazad@c3.ai} \\
  \And
  Henrik Ohlsson \\
  C3 AI \\
  1400 Seaport Blvd\\ 
  Redwood City, CA 94063\\
  \texttt{henrik.ohlsson@c3.ai} \\
}

\begin{document}
\raggedbottom

\maketitle

\begin{abstract}
Traditional time series models are task-specific and often depend on dataset-specific training and extensive feature engineering. While Transformer-based architectures have improved scalability, foundation models, commonplace in text, vision, and audio, remain underexplored for time series and are largely restricted to forecasting.

We introduce \textbf{\texttt{CHARM}}, a foundation embedding model for multivariate time series that learns shared, transferable, and domain-aware representations. To address the unique difficulties of time series foundation learning, \texttt{CHARM} incorporates architectural innovations that integrate channel-level textual descriptions while remaining invariant to channel order. The model is trained using a Joint Embedding Predictive Architecture (JEPA), with novel augmentation schemes and a loss function designed to improve interpretability and training stability. Our $7$M-parameter model achieves state-of-the-art performance across diverse downstream tasks, setting a new benchmark for time series representation learning.
\end{abstract}

\section{Introduction}

Time series models play a pivotal role in critical real-world applications such as forecasting, classification, and anomaly detection across domains including manufacturing, energy, healthcare, and finance \citep{han:19, sus:14, din:15}. By converting temporal signals into actionable insights, these models enable large-scale, data-driven decision-making. However, most existing approaches remain narrowly scoped and task-specific, requiring significant manual effort for development and maintenance. Even in ostensibly homogeneous settings—such as industrial pump fleets with varied sensor configurations—models are often trained independently, despite underlying shared physical dynamics. This fragmentation is rooted in structural limitations of conventional time series architectures, which typically assume fixed-length, uniformly structured inputs and lack mechanisms for fusing information across heterogeneous sensors. Consequently, current paradigms struggle to generalize across tasks, domains, and configurations, posing challenges to scalability and adaptability.

\paragraph{Foundation models in other modalities} In contrast, fields such as natural language processing, computer vision, and audio have undergone transformative progress with the emergence of foundation models—large-scale, pre-trained architectures that learn general-purpose representations across diverse downstream tasks \citep{dev:19, nus:25, ass:23, kir:23, bae:20, bro:20, rad:21}. These models, often trained via Self-Supervised Learning (SSL) on massive unlabeled corpora, have demonstrated capabilities such as Retrieval-Augmented Generation (RAG) \citep{lew:20} and robust task transfer via lightweight fine-tuning \citep{dev:19, oqu:23, kir:23}. Their success hinges on learning semantically meaningful representations that are modular, robust, and highly transferable.

\paragraph{Foundation models for time series forecasting} Inspired by these advances, the time-series community has begun developing foundation models, with a strong emphasis on supervised forecasting objectives \citep{das:24, woo:24, ans:24, liu:24}. These models achieve impressive performance on predictive benchmarks and introduce architectural innovations tailored to multi-domain forecasting. However, because their training remains tightly coupled to a forecasting loss, the learned representations are often specialized and brittle—limiting their applicability to downstream tasks such as classification, segmentation, or anomaly detection. 

\paragraph{Foundation embedding models for time series} Most self-supervised foundation models for time series rely on objectives such as masked reconstruction or next-step forecasting, which require the encoder to impute raw signal values. These signals are often noisy, low-resolution, and entangled with domain-specific artifacts \citep{trirat2024universaltimeseriesrepresentationlearning}, resulting in representations that overfit to sensor-level noise rather than capturing higher-level process dynamics. While such objectives are straightforward to implement, they tend to entangle semantic structure with noise, limiting robustness and generalization across tasks or domains. Recent approaches such as MOMENT (univariate) \citep{goswami2024momentfamilyopentimeseries} and UniTS (multivariate) \citep{gao2024unitsunifiedmultitasktime} extend this paradigm with multi-task or reconstruction-based pretraining and report strong downstream performance, but remain fundamentally grounded in raw signal-level prediction for pretraining.

\paragraph{JEPA-style latent prediction: a robust alternative} In contrast, Joint Embedding Predictive Architectures (JEPA) \citep{lec:22} adopt a fundamentally different approach: predicting latent representations of masked target segments from contextual embeddings rather than raw values. By operating entirely in embedding space, JEPA filters out sensor noise and encourages the encoder to model higher-level temporal structure. In vision, this paradigm has proven highly effective—\citet{ass:23} demonstrate that latent prediction yields representations that rival or surpass those learned via supervised learning, while remaining more robust to noise and label scarcity. Compared to contrastive learning, JEPA-style models also avoid the complexity of negative sampling and the sensitivity to embedding space dimensionality, making them a more stable and scalable choice for semantic representation learning.

\paragraph{Lack of channel-awareness in time series models} Most time series models treat input channels as undifferentiated numerical streams, without incorporating information about the identity, modality, or semantics of the sensors generating the data. This lack of sensor-awareness discards valuable contextual information, limiting the model’s ability to reason about sensor-specific behavior or operate reliably across deployments with varying instrumentation.

\subsection{Contributions}

This work aims to (1) develop a robust, semantically grounded SSL objective for time series data, and (2) design an architecture capable of directly incorporating textual channel information. This is inspired by how subject matter experts interpret time series data, by jointly considering the raw signals and their accompanying channel descriptions. To this end, we introduce a   \textbf{CH}annel-\textbf{A}ware \textbf{R}epresentation \textbf{M}odel (\texttt{CHARM}), trained to produce domain-aware, and performant representations across tasks and datasets. Building such a model entails several key challenges, including channel heterogeneity, variation in temporal dynamics across domains, and risks of negative transfer and representational collapse. To realize these aims, we introduce the following core contributions:

\paragraph{Description-aware temporal featurization} We modify temporal convolutional networks to incorporate channel descriptions directly into the convolutional layers. Unlike patch-based approaches, our stacked, description-aware convolutions allow the model to seamlessly adapt across domains without manual tuning of patch size. Details are provided in Section~\ref{sec:ctcn}.
    
\paragraph{Inter-channel reasoning via attention and gating} We augment the standard attention mechanism with novel, learnable inter-channel attention layers and gating modules conditioned on channel descriptions. These components enable the model to flexibly capture inter-channel dependencies, selectively integrate signals in a structured manner, while maintaining invariance to channel ordering. See Section~\ref{sec:cal} for details.

\paragraph{Self-supervised training with JEPA for time series} We adapt the JEPA to the time series domain, enabling semantic representation learning without reconstruction. To do so, we introduce a set of tailored data augmentations and temporal perturbations that improve robustness to common time series artifacts. This avoids the drawbacks of contrastive learning, such as sensitivity to sampling and dimensionality constraints~\citep{lec:22, ass:23, che:20, che:22, chu:20}. Details are provided in Section~\ref{sec:SSL}.
    
%\paragraph{A unified loss for stable representation learning} We design a custom loss that combines self-supervised objectives with architectural regularization to support efficient and stable learning. See Section~\ref{sec:loss} for more details.

We evaluate our model across several downstream classification, forecasting, and anomaly detection tasks, where it achieves state-of-the-art performance, often outperforming specialized, task-specific models.

\section{Methodology}
In this section, we first introduce a novel multi-modal transformer-based architecture for learning embeddings from time series data, guided by underlying channel descriptions (\Cref{sec:arch}). We then describe how this architecture is trained using self-supervised learning with JEPA (\Cref{sec:SSL}). The notation used throughout this section is provided in Appendix~\ref{app:notation}.

\subsection{Multi-Modal Time Series Embedding Model} \label{sec:arch}
\begin{figure}
    \centering
    \includegraphics[width=0.9\linewidth]{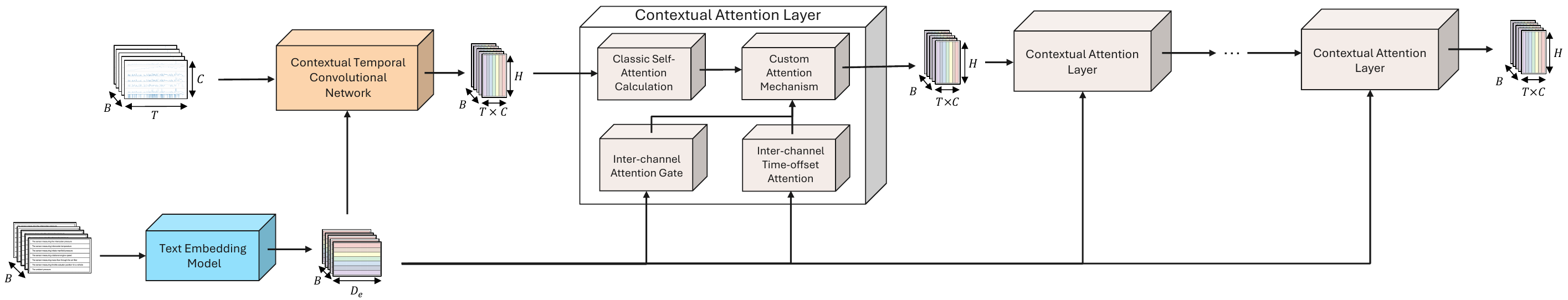}
    \caption{Overview of the model architecture, featuring a context-aware temporal convolutional network and a series of contextual attention layers, each guided by textual descriptions of the input time series channels.}
    \label{fig:architecture}
\end{figure}

Here, we present three key architectural contributions that enable learning high-quality time series embeddings by incorporating textual channel descriptions. Our model employs convolutional layers in conjunction with a series of custom attention layers, enhanced by a novel attention mechanism. An overview of the full architecture is provided in \Cref{fig:architecture}.

We begin by describing the contextual temporal convolutional network in \Cref{sec:ctcn}, which generates convolution-based embeddings. These embeddings are then passed to a series of contextual attention layers, where our novel attention mechanism is applied. We describe the details of this layer in \Cref{sec:cal}, where we introduce two core extensions to the self-attention mechanism in sections.

\subsubsection{Contextual Temporal Convolutional Network} \label{sec:ctcn}

\begin{figure}
    \centering
    \includegraphics[width=0.9\linewidth]{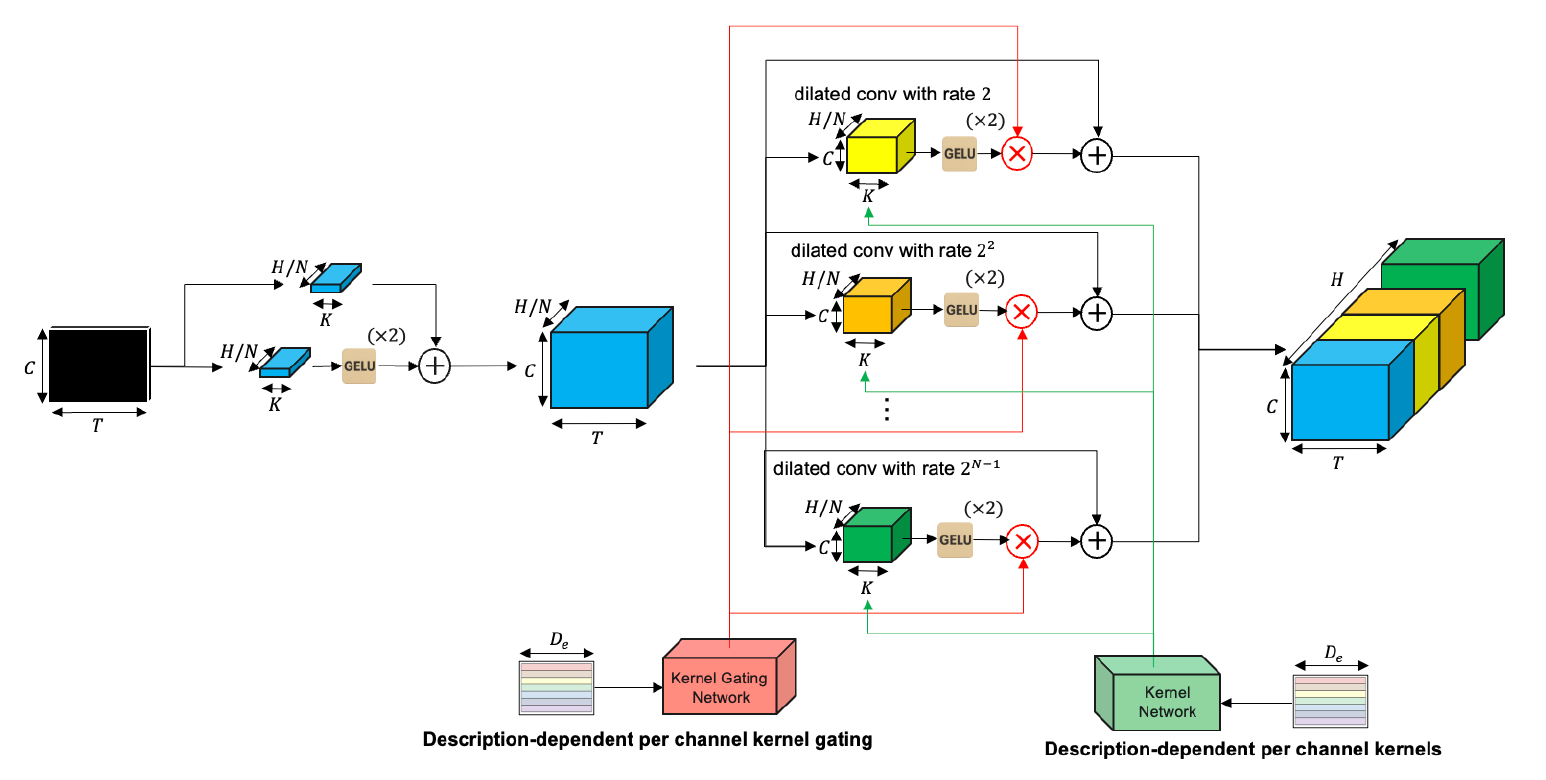}
    \caption{Schematic of the context-aware temporal convolutional network, performing initial featurization of multivariate time series inputs guided by granular textual descriptions of each channel.}
    \label{fig:tcn}
\end{figure}

We introduce a contextual Temporal Convolutional Network (TCN)  that projects input time series $\mathbf{T} \in \mathbb{R}^{T \times C}$ into contextual embeddings $\mathbf{T}_c \in \mathbb{R}^{T \times C \times H}$, where $\mathbf{T}_c[i,j,:]$ denotes the $H$-dimensional embedding at time step $i$ for channel $j$. The base architecture follows standard dilated TCNs~\citep{bai:18, lin:21}, which stack $1$D convolutions with exponentially increasing dilation factors ($2^{l}$). However, standard TCNs are architecturally static and their learned kernels are input independent and constant. This lack of flexibility hinders their ability to adapt across diverse domains, leading to representation collapse or negative transfer when trained on heterogeneous datasets. To address this, we make the TCN \textit{context-aware} by incorporating channel descriptions into the convolutional process. Given a time series tuple $\mathbf{t} = (\mathbf{T}, \mathbf{D}, \mathbf{pos})$, we extract text embeddings for the descriptions using a frozen text embedding model, as $\mathbf{E}_d \in \mathbb{R}^{C \times D_e}$. We introduce two mechanisms to inject this context, namely:

\paragraph{Contextual kernel gating}\label{channel_gates_desc} Description embeddings are used to conduct soft gating through the layers of the TCN. The gates are produced by the \emph{kernel gating network} in \Cref{fig:tcn}, which is given as
$\mathbf{G}_c = \textbf{sigmoid}(\mathbf{E}_d \mathbf{W}_g), \mathbf{W}_g \in \mathbb{R}^{D_e \times N},$
with $N$ denoting the number of stacked convolutional layers in the TCN. Each element $\mathbf{G}_c[i,j]$, which corresponds to the soft gate associated with channel $i$ and layer $j$ of the TCN which is then incorporated multiplicatively in the network as depicted in \Cref{fig:tcn}. This enables the model to control the effective field of view of TCN informed by the channel descriptions. 

\paragraph{Contextual kernels} Rather than learning fixed convolutional filters, we generate them from the descriptions embeddings as
$\mathbf{G}_k = \mathbf{E}_d \mathbf{W}_k,  \mathbf{W}_k \in \mathbb{R}^{D_e \times (\frac{H\times K}{N})},$
where $K$ is the kernel size and $N$ the number of TCN layers. This mechanism directly ties channel semantics to filter generation and is represented by the \emph{kernel network} in \Cref{fig:tcn}.

\subsubsection{Contextual Attention Layer} \label{sec:cal}
The embeddings generated by the contextual TCN layer are subsequently processed through a sequence of contextual attention layers. The primary goal of these layers is to effectively fuse channel and temporal dimensions into richer, more expressive representations, directly incorporating the granular textual descriptions of each channel. To achieve this, we propose several novel extensions to the classical self-attention mechanism~\citep{vas:17}. These and their inter-play are depicted in \Cref{fig:architecture}, under the contextual attention layer. Below we discuss the key details of these key components in detail.

\begin{figure}[htbp]
    \centering
    \begin{minipage}[b]{0.4\textwidth}
        \centering
        \includegraphics[width=\textwidth]{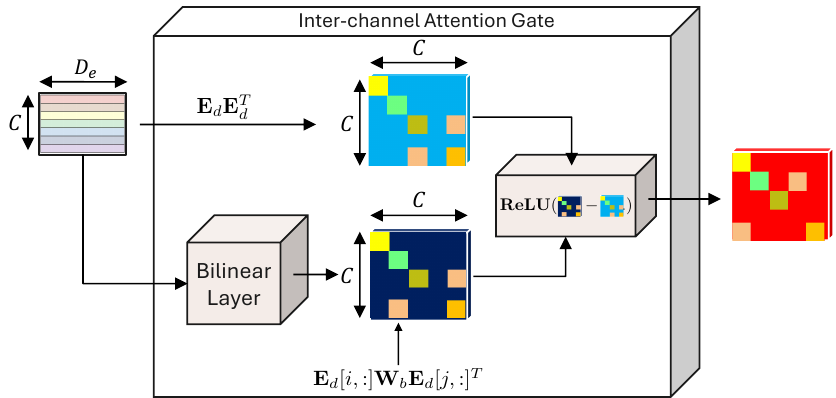}
        \caption{Description-aware gating mechanism, selectively suppressing cross-channel attention.}
        \label{fig:channel_gate}
    \end{minipage}
    \hfill
    \begin{minipage}[b]{0.5\textwidth}
        \centering
        \includegraphics[width=\textwidth]{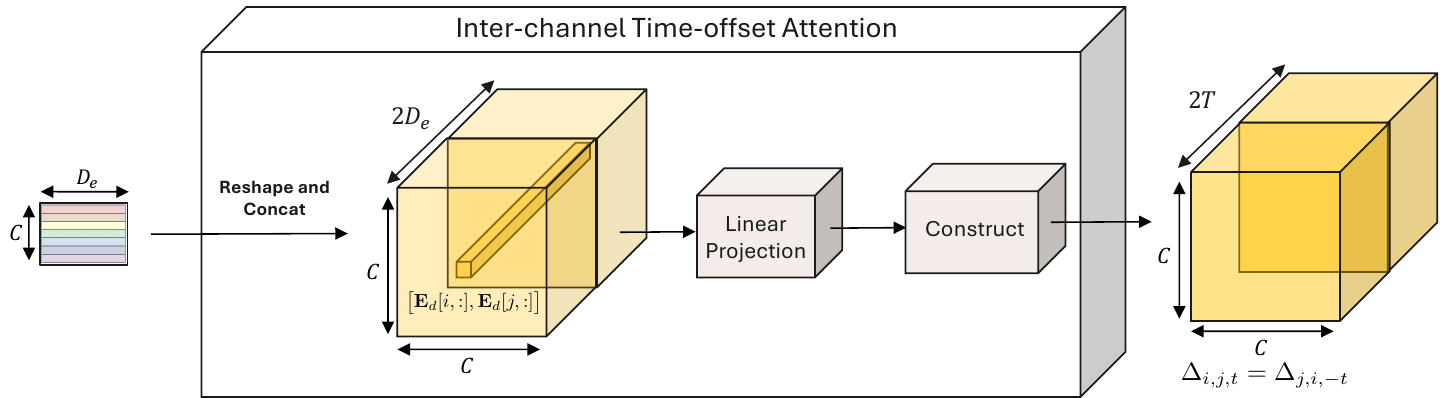}
        \caption{Symmetric construction of inter-channel temporal-offset attention, encoding mutual dependencies between channels at varying time lags.}
        \label{fig:time_dalta}
    \end{minipage}
\end{figure}

\paragraph{Description-aware inter-channel attention gating}
\label{para:channel_gates}
This module introduces a gating mechanism conditioned explicitly on channel descriptions, enabling our architecture to selectively co-attend to the most relevant channels. Given the channel description embeddings $\mathbf{E}_d$, this layer computes the pairwise similarities, $\mathbf S$, and the similarity threshold matrix, $\mathbf{Z}$, as 
$\mathbf{S} = \mathbf{E}_d \mathbf{E}_d^\top, \quad
\mathbf{Z}[i, j] = \text{sigmoid}\left(\mathbf{E}_d[i,:]\,\mathbf{W}_b\,\mathbf{E}_d[j,:]^\top\right)$. The similarity threshold matrix governs our inter-channel gating mechanism. Specifically, this layer outputs the gating matrix given as
$\mathbf{G}_d = \text{ReLU}(\mathbf{Z} - \mathbf{S}).$
This process, illustrated in \Cref{fig:channel_gate}, allows the model to selectively suppress cross-attention between channel pairs $(i, j)$ by driving their corresponding similarity threshold $\mathbf{Z}[i, j]$ toward one.  

\paragraph{Description-aware inter-channel time-offset attention} 
This module improves the model's ability to capture inter-channel relationships by explicitly quantifying dependencies between channels at different temporal offsets. Specifically, we introduce a learnable tensor $\mathbf{\Delta}\in\mathbb{R}^{C\times C\times 2T_{\text{max}}}$, where each entry $\Delta_{i,j,t}$ encodes the learned dependency strength between channel $i$ and channel $j$ at a temporal offset of $t$ steps. We assume inherent symmetry within $\mathbf{\Delta}$, reflecting the intuition that the relationship from channel $i$ to channel $j$ at step $t$ should match the inverse relationship from channel $j$ to channel $i$ at step $-t$, formally as $\mathbf{\Delta}_{i, j, t} = \mathbf{\Delta}_{j, i, -t}$.
To explicitly enforce this symmetry, we follow a structured construction procedure. Given the channel description embeddings $\mathbf{E}_d$, we first create a pairwise embedding tensor $\bar{\mathbf{E}}_d \in \mathbb{R}^{C\times C\times 2D_e}$ defined by concatenation as
\[
\bar{\mathbf{E}}_d[i, j, :] = [\mathbf{E}_d[i, :], \mathbf{E}_d[j, :]] \in \mathbb{R}^{2D_e}.
\]
Next, we apply a linear projection to these pairwise embeddings, parameterized by the matrix $\mathbf{W}_d \in \mathbb{R}^{2D_e \times T_{\text{max}}}$, yielding the intermediate tensor $\mathbf{\Delta}_{+} = \bar{\mathbf{E}}_d\,\mathbf{W}_d$. We then construct the full symmetric tensor $\mathbf{\Delta}$ as below
\[
\mathbf{\Delta}[i, j, t]=
\begin{cases}
    \mathbf{\Delta}_{+}[i, j, t] & \text{if } t \geq 0, \\[6pt]
    \mathbf{\Delta}_{+}[j, i, -t] & \text{if } t < 0.
\end{cases}
\]

This symmetric construction, depicted in \Cref{fig:time_dalta}, ensures parameter efficiency and explicitly encodes symmetry constraints. We compute the final $\bar{\mathbf{\Delta}}\in \mathbb{R}^{CT\times CT}$ matrix using a "slice-and-tile" operation, where $\bar{\mathbf{\Delta}}$ is a block matrix with $T$ blocks on each axis, and in block notation  $\bar{\mathbf{\Delta}}[T_i, T_j]=\mathbf{\Delta}[:,:,T_j-T_i]$. See \Cref{app: fast_attn} for PyTorch style pseudocode for the naive and fast versions of this operation.

\paragraph{Custom attention mechanism}
We unify the gating and attention mechanisms described above into a single self-attention framework. Given embedding matrix the contextual TCN layer, $\mathbf{T}_c$, we reshape it into $\mathbf{X} \in \mathbb{R}^{CT \times H}$, where each channel-time pair is represented by an $H$-dimensional embedding. To facilitate intuitive indexing, we employ a triple-index notation $\mathbf{X}_{[(c_i,t_j),k]}$ rather than a flattened indexing scheme $\mathbf{X}[m, k]$, with $c_i = m \mod C$ and $t_j = \lfloor \frac{m}{C} \rfloor$. First we apply rotary position embeddings to the queries and keys given the $\mathbf{pos}$ indices as:
\begin{align*}
    \mathbf{\hat{Q}} = \mathbf{RoPE}(\mathbf{W}_{Q}\mathbf{X}_{[(i,p),:]}, \mathbf{pos}), \quad
    \mathbf{\hat{K}} = \mathbf{RoPE}(\mathbf{W}_{K}\mathbf{X}_{[(j,q),:]}, \mathbf{pos}) 
\end{align*}
The custom attention matrix $\mathbf{A} \in \mathbb{R}^{CT \times CT}$ is then constructed as
\begin{align*}
    \mathbf{A}_{[(i,p),(j,q)]} = \text{Softmax}\left( 
        \underbrace{\frac{\mathbf{\hat{Q}}_{[i,p,:]}\mathbf{\hat{K}}_{[j,q,:]}^T}{\sqrt{D_e}}}_{\text{Vanilla Self-Attention}}
        + \underbrace{\mathbf{\Delta}[i,j,q-p]}_{\text{Channel Lags}}
        - \underbrace{\lambda_G \mathbf{G}_d[i,j]}_{\text{Channel Gates}}
    \right)
\end{align*}
Here, $\mathbf{A}_{[(i,p),(j,q)]}$ represents the attention from channel $i$ at time $p$ to channel $j$ at time $q$. The scalar $\lambda_G$ is typically a large positive number, enabling the gating matrix to serve as an attention mask, selectively blocking certain cross-channel interactions based on the learnt thresholds. The attention matrix can be efficiently computed using vectorized operations by appropriately tiling the inter-channel gating and time-offset matrices. Following the standard transformer approach, we multiply the attention matrix by the value matrix $\mathbf{V} = \mathbf{W}_V\mathbf{X}$ to produce our contextualized embeddings.

\subsubsection{Putting It All Together}

This completes the integration of the various components within our multimodal time-series embedding architecture. For a given input tuple $\mathbf{t} = (\mathbf{T}, \mathbf{D}, \mathbf{pos})$, we first generate the initial embeddings $\mathbf{X}\in\mathbb{R}^{T\times C \times H}$ from our \textbf{contextual TCN} layer. These embeddings pass through a stack of $N$ \textbf{contextual attention} layers, each layer outputting $\mathbf{X}^{(l)} \in \mathbb{R}^{(T \times C) \times H}$, reshaped to $\mathbf{Y}^{(l)} \in \mathbb{R}^{T \times C \times H}$ for subsequent layers. Similar to \citep{gri:20}, we apply $\ell_2$ normalization to the final embeddings:
\[
\mathbf{Y}[i, j, t] = \frac{\mathbf{Y}[i, j, t]}{\sqrt{\sum_{h}\mathbf{Y}[i, j, h]^2}},
\]
with normalization computed along the embedding dimension only. The complete architecture is denoted as $\mathbf{E}_\theta$, such that $\mathbf{Y}=\mathbf{E}_\theta(\mathbf{T},\mathbf{D}, \mathbf{pos})$. While this outlines the primary structure of our contextual embedding model, we have also introduced several nuanced modifications aimed at enhancing training stability and convergence speed. These detailed adjustments are presented in \Cref{app: impl_deets}.

\subsection{Self-supervised Representation Learning} \label{sec:SSL}

We adopt the JEPA framework \citep{ass:23, lec:22} to enable self-supervised learning on time series data enriched with fine-grained textual context. JEPA comprises three core components, namely predictor, context, and target encoders. In the following section, we detail the key components of our training pipeline based on JEPA, namely, (i) the dataset generation process in \Cref{sec:dataset_gen}, (ii) the integration of JEPA with our embedding model in \Cref{sec:jepa_setup}, and (iii) a novel loss formulation tailored to JEPA training in \Cref{sec:loss}. 

\subsubsection{Dataset Generation}\label{sec:dataset_gen}
\begin{figure}
    \centering
    \includegraphics[width=0.45\linewidth]{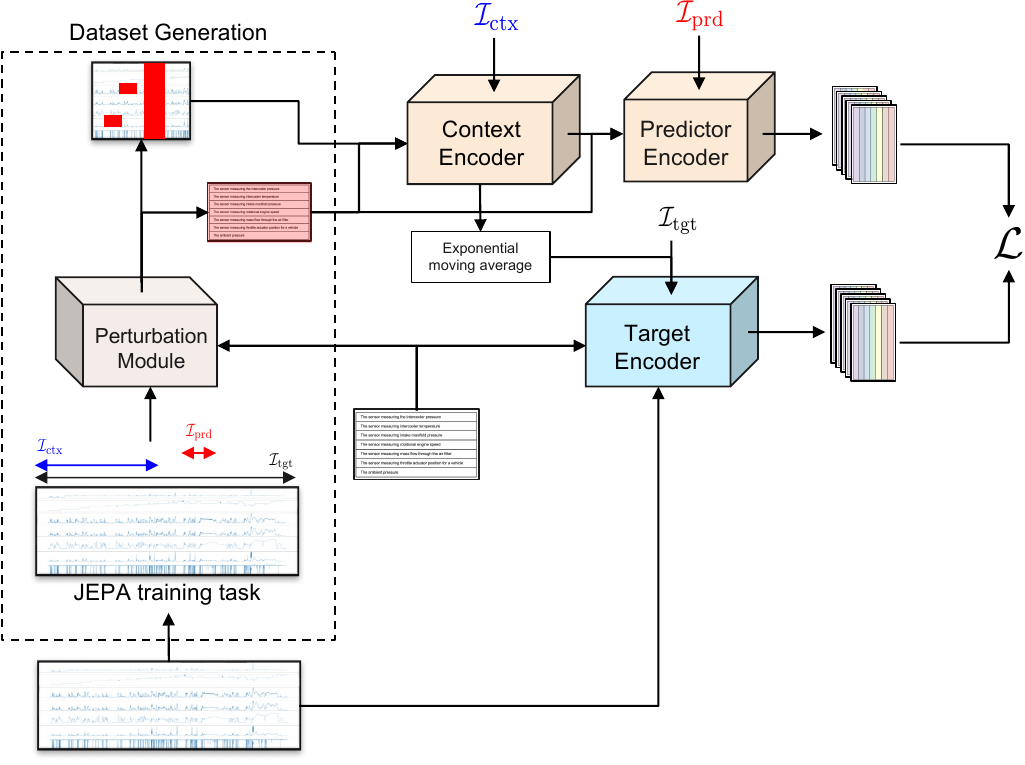}
    \caption{Overview of JEPA, showing its three encoders and how they process augmented views of the same data instance.}
    \label{fig:jepa}
\end{figure}

\Cref{fig:jepa} provides a high-level view of JEPA and the interplay among JEPA's three encoders, and how they consume data points generated by the dataset generation process. For time series data, we generate augmented views of the same data point through data augmentation and perturbation techniques. We refer to the data augmentation as JEPA training tasks. 

Formally, given an input instance $\mathbf{t}=(\mathbf{T}, \mathbf{D}, \mathbf{pos})$, we define an augmented view of this instance through three randomly generated contiguous sets of indices, with $\mathcal{I}_{\text{ctx}}, \mathcal{I}_{\text{tgt}}, \mathcal{I}_{\text{prd}} \subseteq \mathbf{pos}$, denoting time indices that are fed into the context, target and predictor encoders. We employ two self-supervised tasks, namely \emph{causal prediction}, where $\mathcal{I}_{\text{ctx}} \subset \mathcal{I}_{\text{tgt}}$, and \emph{smoothing}, where $\mathcal{I}_{\text{ctx}} \cap \mathcal{I}_{\text{tgt}} \neq \emptyset$ and $\mathcal{I}_{\text{prd}} \subset \mathcal{I}_{\text{ctx}} \cap \mathcal{I}_{\text{tgt}}$. See \Cref{fig:jepa_tasks} for an overview of the causal prediction (left) and smoothing (right) tasks. 

We further perturb the input of the context encoder, while keeping the input to the target encoder clean and unaltered. This design encourages the model to learn robust representations despite corrupted or noisy context. Specifically, motivated by artifacts commonly encountered in real-world time series data, we mask random segments of the time series data, either uniformly across all channels or selectively across a subset of randomly chosen channels. This is designed to enhance the model's robustness to practical data imperfections and improve generalization in downstream tasks.

\subsubsection{JEPA Setup}\label{sec:jepa_setup}
In JEPA the context and target encoders are architecturally identical. However, only the context encoder is directly optimized during training while the target encoder is updated using an exponential moving average of the context encoder's parameters, see \Cref{fig:jepa}. In contrast, the predictor is a narrow/shallower version of the context encoder which is trained jointly with the context encoder through standard backpropagation.  

To leverage JEPA, we integrate our embedding model within the underlying encoders. Let us denote the context, target and predictor encoders as, $\mathbf{E}^c_\theta, \mathbf{E}^t_\theta$ and $\mathbf{E}^p_\theta$, respectively. The context and target encoders are fully defined by our embedding model as 
\begin{align*}
\mathbf{X}^c = \mathbf{E}_\theta^c(\bar{\mathbf{T}}, \bar{\mathbf{D}},\mathcal I_{\text{ctx}}) \coloneqq \mathbf E_\theta (\bar{\mathbf{T}}, \bar{\mathbf{D}},\mathcal I_{\text{ctx}}), \quad
\mathbf{X}^t = \mathbf{E}_\theta^t(\mathbf{T}[\mathcal I_{\text{tgt}}, :], \mathbf{D},\mathcal I_{\text{tgt}}) \coloneqq \mathbf E_\theta ([\mathcal I_{\text{tgt}}, :], \mathbf{D},\mathcal I_{\text{tgt}})
\end{align*}
where $\mathbf{X}^c \in \mathbb{R}^{|\mathcal I_{\text{ctx}}|\times C\times H}$ and $\mathbf{X}^t \in \mathbb{R}^{|\mathcal I_{\text{tgt}}|\times C\times H}$ are the outputs from the context and target encoders, respectively, and $\bar{\mathbf{D}}$ and $\bar{\mathbf{T}} \in \mathbb{R}^{|\mathcal I_{\text{ctx}}| \times C}$ denote the perturbed descriptions and the perturbed time series data $\mathbf T[\mathcal I_{\text{ctx}}, :]$. The predictor encoder accepts the output of the context encoder as input. Unlike the context and target encoders, the predictor encoder solely leverages the contextual attention layer. Let $\bar{\mathbf{A}}_\theta$ denote the narrower and shallower version of the contextual attention layer. Also let 
\[
\bar{\mathbf X}_c = [ \mathbf X^c, \underbrace{\mathbf m_\theta, \cdots, \mathbf m_\theta}_{\text{repeated $|\mathcal I_{\text{prd}}|$ times}} ]
\]
where $\mathbf m_\theta$ represents learnable placeholders that guide the predictor encoder to generate embeddings for masked positions, see \citep{ass:23} for more information. We further define the concatenated set $\bar{\mathcal I}_{\text{prd}} = \mathcal I_{\text{ctx}} + \mathcal I_{\text{prd}}$. The predictor encoder is then defined as $\mathbf{X}^p = \mathbf{E}_\theta^p(\bar{\mathbf X}^c, \bar{\mathbf{D}}, \bar{\mathcal I}_{\text{prd}}) \coloneqq  \bar{\mathbf{A}}_\theta (\bar{\mathbf X}^c\mathbf{W}_{pd}, \bar{\mathbf{D}}, \bar{\mathcal I}_{\text{prd}}) \mathbf{W}_{pu},$
where $\mathbf{X}^p \in \mathbb{R}^{|\bar{\mathcal I}_{\text{prd}}| \times C \times H}$ is the output of the predictor encoder, and $\mathbf{W}_{pu} \in \mathbb{R}^{H_d \times H}$,  $\mathbf{W}_{pd} \in \mathbb{R}^{H \times H_d}$ denote linear layers used for up and down projecting. 

\subsubsection{Training Loss}\label{sec:loss}
The training loss for training our embedding model comprises two major components, self-supervised objectives and regularization terms associated with key modules of the contextual attention layer.

\paragraph{Self-supervised loss} Our embedding model produces embeddings at the level of each time point and each channel. We employ a self-supervised objective based on the $\ell_1$ norm to measure discrepancies between embeddings from two augmented views of the same time series instance. To promote consistency not only at the most granular level but also across coarser aggregations, we extend the objective to include progressively aggregated embeddings. Let $\bar{\mathbf X}^t = \mathbf X^t[\mathcal I_{\text{prd}}, :] \in \mathbb{R}^{|\mathcal I_{\text{prd}}| \times C \times H}$ and $\bar{\mathbf X}^p = \mathbf X^p[-|\mathcal I_{\text{prd}}|:, :] \in \mathbb{R}^{|\mathcal I_{\text{prd}}| \times C \times H}$. The self-supervised loss is then defined as
\begin{align*}
    \mathcal{L}_{\text{ssl}} 
    &= \sum_{i,j,t} \left| \bar{\mathbf{X}}^p_{i,j,t} - \bar{\mathbf{X}}^t_{i,j,t} \right|
    +\sum_{i,t} \left| \mu_j\left( \bar{\mathbf{X}}^p_{i,:,t} \right) - \mu_j\left( \bar{\mathbf{X}}^t_{i,:,t} \right) \right| 
    +\sum_{t} \left| \mu_{i,j}\left( \bar{\mathbf{X}}^p_{:,:,t} \right) - \mu_{i,j}\left( \bar{\mathbf{X}}^t_{:,:,t} \right) \right|
\end{align*}
with $\mu_j(\mathbf{X}_{i,:,t}) = \frac{1}{C} \sum_j \mathbf{X}_{i,j,t}, \quad
\mu_{i,j}(\mathbf{X}_{:,:,t}) = \frac{1}{CT} \sum_i \sum_j\mathbf{X}_{i,j,t}$. This multi-resolution loss encourages the model to align representations both at the fine-grained level (per time point and channel) and at higher levels of abstraction (per time point and globally), thereby enhancing regularity and usability of the embeddings at different levels of granularity.
\paragraph{Regularization loss} We include two regularization terms related to key modules of the contextual attention layer, namely the inter-channel gating and inter-channel time-offset attention modules. Given the inherent sparsity in meaningful channel relationships, we promote sparsity in the learned channel relationships, by regularizing the similarity threshold matrix $\mathbf{Z}$ toward $1$ and regularizing the relationships among channels across temporal offsets using
\[
R_1 = \sum_{i,j} \left|1 - \mathbf{Z}[i,j]\right|, \quad R_2 = \frac{\sum_{i,j}\sum_{t}\mathbf{\Delta}[i, j, t]^2}{C^2},
\]
respectively. The regularization term $R2$ encourages consistency and stability in the learned inter-channel temporal relationships. Combining these loss terms results in the following training objective function to be applied across all data points  $\mathcal L = \mathcal L_{\text{ssl}} + \lambda_1 R_1 + \lambda_2 R_2,$ where $\lambda_1$ and $\lambda_2$ control the extent and strictness of gating and temporal attention suppression.

\section{Experiments}
In this section, we evaluate our model's embeddings on common downstream tasks, namely classification, forecasting, and anomaly detection, and benchmark our model against the current state-of-the-art models in each of the aforementioned downstream tasks. We expand on our datasets \Cref{app:datasets}, training, and baselines in \Cref{app: experiments}.

\newcommand{\supera}{\textsuperscript{$\alpha$}}
\newcommand{\superb}{\textsuperscript{$\beta$}}
\newcommand{\superc}{\textsuperscript{$\gamma$}}

% \paragraph{Baselines} We compare our model's performance against baseline scores reported by the following categories of models:
% \begin{itemize}[leftmargin=1.5em]
%     \item \textbf{Representation Learning Based Models}\supera{}: (T-REP, TS2Vec, T-Loss etc.)
    
%     These models are categorized on the basis that they are truly unsupervised, and use no task specific information to learn improve their representations. Our model falls most naturally into this category, as we follow the: frozen encoder $\rightarrow$ probing protocol, similar to the other models in this category.
    
%     \item \textbf{Forecasting Based Models}\superb{} : (AutoFormer, FEDFormer, Informer, etc.) \\\textbf{(only for forecasting tasks)}
    
%     These models are trained on specific datasets, purely for forecasting and do not output general purpose representations.
    
%     \item \textbf{Supervised Task Models}\superc{} : (UniTS, MOMENT etc.)

%     These models learn representations in a supervised manner, i.e. the downstream task is used for training the model. However, these models tend to be universal, in that they can be used with multiple datasets and tasks without re-training.
    
% \end{itemize}

% Due to compute restrictions, we report the baseline scores and evaluation protocols as outlined in the original papers directly.

\captionsetup[subtable]{%
  position=bottom,   % or 'top' if you want captions above
  justification=centering,
  skip=1ex           % adjust vertical space between table and caption
}
\begin{table*}[!ht]
  \centering
  \footnotesize
  \begin{subtable}[c]{0.45\textwidth}   % change [t] → [c]
    \centering
    \begin{tabular}{lccc}
      \toprule
      \textbf{Method} & \textbf{F1}↑  & \textbf{FAR}(\%)↓ & \textbf{MAR}(\%)↓ \\
      \midrule
      Conv‑AE              & 0.78 & 13.55 & 28.02 \\
      MSET                 & 0.78 & 39.73 & 14.13 \\
      T‑squared+Q (PCA)    & 0.76 & 26.62 & 24.92 \\
      Isolation Forest     & 0.29 & \textbf{2.56}  & 82.89 \\
      LSTM-VAE             & 0.56 & 9.13  & 55.03 \\
      MSCRED               & 0.36 & 49.94 & 69.88 \\
      \rowcolor{lightblue}
      \textbf{CHARM}        & \textbf{0.86} & 19.35 & \textbf{12.69} \\
      \bottomrule
    \end{tabular}
    \caption{\textbf{SKAB Anomaly Detection Benchmark}\footnotemark[1]\\
    \parbox[t]{\linewidth}{
    \tiny
    The baselines use autoencoder based reconstruction techniques, and are exclusively trained on the benchmark dataset directly; our model is frozen with a lightweight reconstruction head on top. See \Cref{app:anomaly_detection} for details.
    }
    }
    \label{tab:anomaly}
  \end{subtable}%
\hfill
  \begin{subtable}[c]{0.45\textwidth}   % change [t] → [c]
    \centering
    \begin{tabular}{lcc}
      \toprule
      \textbf{Method}      &\textbf{ Avg.~Acc.}~↑ & \textbf{Ranks 1st}↑ \\
      \midrule
      T‑Loss  & 0.712 &  5 \\
      TS2Vec  & 0.769 &  2 \\
      TNC    & 0.721 &  1 \\
      TS‑TCC & 0.731 &  2 \\
      DTW                         & 0.694 &  3 \\
      T‑Rep   & 0.773 &  1 \\
      \rowcolor{lightblue}
      \textbf{CHARM}    & \textbf{0.788} & \textbf{7} \\
      \bottomrule
    \end{tabular}
    \caption{\textbf{UEA classification}\footnotemark[2] \\
    \parbox[t]{\linewidth}{
    \tiny
    The baselines are exclusively trained on each dataset separately; our model is frozen with an SVM classifier to probe the representations. See \Cref{app: classification} for details.
    }
    }
    \label{tab:uea}
  \end{subtable}
  \caption{Comparison on anomaly detection and classification.}
  \label{tab:combined}
\end{table*}

\footnotetext[1]{Reported scores from official SKAB benchmark\citep{skab}}
\footnotetext[2]{Reported scores from T-Rep \citep{fra:24}}

\paragraph{Classification}

To evaluate the discriminative power of our learned representations, we benchmark our model on the widely-used UEA multivariate time series classification suite \citep{uea}.

As seen in \Cref{tab:uea} our model establishes a new state-of-the-art on the UEA multivariate classification benchmark among representation learning-based methods, exceeding prior performance by \textbf{1.5\%} on average. It also attains the highest number of \textbf{wins}, demonstrating superior generalization across diverse datasets. Detailed per-dataset scores, evaluation protocols, and statistical comparisons are reported in the \Cref{app: classification}.

\paragraph{Anomaly detection}

Anomaly detection revolves around predicting abnormal behaviors in time series data, and is hugely relevant in industrial use-cases where optimal/stable operation of machinery is critical to safety.
To assess our model's ability to detect such abnormal behavior in time series, we evaluate our model's performance on the Skoltech Anomaly Benchmark (SKAB), a popular multivariate anomaly detection benchmark, where anomalies are classified based on reconstruction errors between the expected and true behavior of a sensor (see \Cref{app:anomaly_detection} for more details regarding the SKAB benchmark, and our linear probing setup). 

\Cref{tab:anomaly} summarizes our model's performance on the SKAB benchmark, where we improve the state-of-the-art \texttt{F1} score and \textbf{Missed Alarm Rate (MAR)} by \textbf{0.08} and \textbf{~$\mathbf{10}\%$} respectively.

\paragraph{Forecasting}
We evaluate our model’s ability to capture temporal dependencies and generalize to future time points through standard time series forecasting tasks. We quantitatively assess the model's performance using mean squared error (MSE) and mean absolute error (MAE) metrics averaged over all forecasted timesteps and across all target variables. We report results across multiple benchmark datasets spanning diverse domains (e.g., electricity, weather, exchange rate, illness etc.), each with varying temporal dynamics and seasonality. 
Full implementation, results, task and dataset details are provided in the Appendix.

As seen in \Cref{fig:comparison_mse} and \Cref{app:forecasting}, our model surpasses the SOTA on the \texttt{ETTh1}, \texttt{ETTh2}, \texttt{ETTm1}, \texttt{ETTm2}, \texttt{Exchange Rate}, \texttt{Illness} datasets, and is competitive on the \texttt{Weather} dataset. We observe a significant improvement on the \texttt{ETT} benchmarks, while achieving near-SOTA scores for all other tasks.
\begin{figure}[ht]
    \centering
     \includegraphics[width=0.24\linewidth]{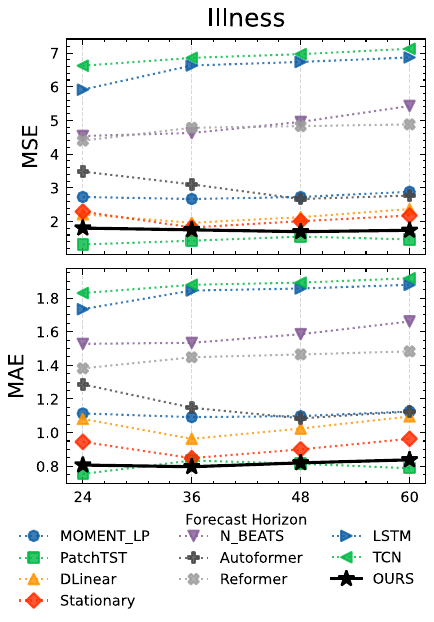}
    \includegraphics[width=0.24\linewidth]{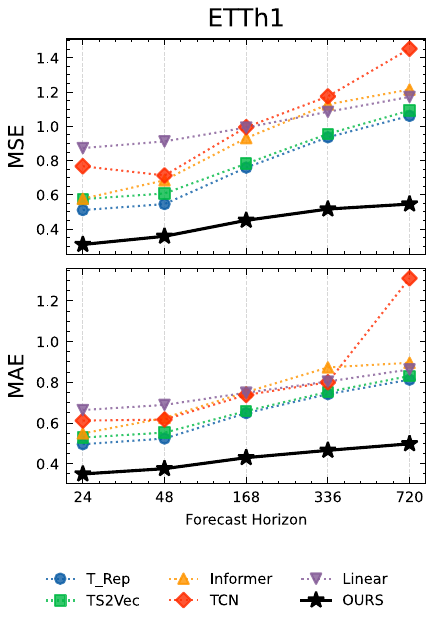}
    \includegraphics[width=0.24\linewidth]{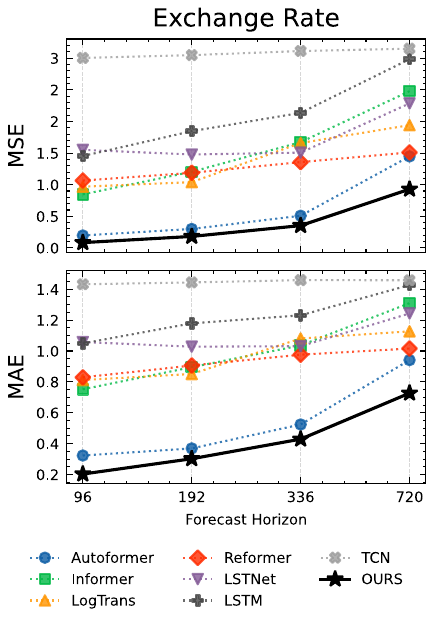}
    \includegraphics[width=0.24\linewidth]{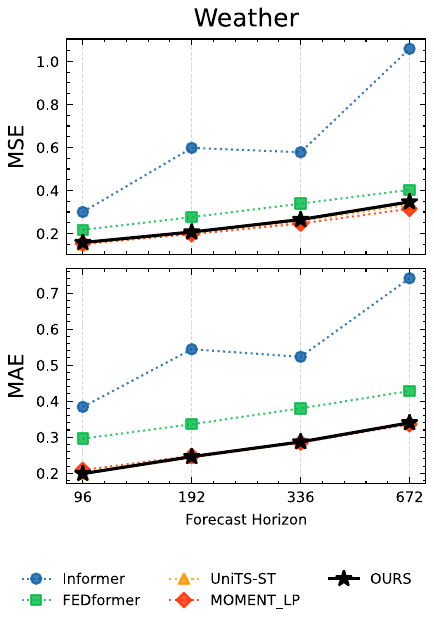}
    \caption{Comparison of our model's errors (MSE, MAE) vs baselines for 4/7 forecasting datasets}
    \label{fig:comparison_mse}
\end{figure}
\begin{table}[ht]
  \centering
  \scriptsize
  \setlength{\tabcolsep}{3pt}
  % first minipage at 45% of text width
  \begin{minipage}[t]{0.48\textwidth}
    \centering
    \begin{tabular}{llccc}
      \toprule
      \textbf{Dataset} & \textbf{Metric}
        & \textbf{Best Baseline (Value)}
        & \textbf{CHARM}
        & \textbf{Absolute $\Delta$} \\
      \midrule
      \rowcolor{lightblue}
      \texttt{ETT (Avg.)} & MSE & T‑Rep (0.986) & 0.333 & \textbf{-0.653} \\
      \rowcolor{lightblue}
       \texttt{ETT (Avg.)} & MAE & T‑Rep (0.702) & 0.358 & \textbf{-0.344} \\
        
      \midrule
        \texttt{Weather} & MSE & MOMENT$_\text{LP}$ (0.228) & 0.244 & +0.016 \\
        \texttt{Weather} & MAE & UniTS‑ST (0.266)         & 0.268 & +0.002 \\
       
      \midrule
      \rowcolor{lightblue}
      \texttt{Exchange Rate}  & MSE & Autoformer (0.613) & 0.387 & \textbf{-0.226} \\
      \rowcolor{lightblue}
      \texttt{Exchange Rate} & MAE & Autoformer (0.539) & 0.415 & \textbf{-0.124} \\
        
      \midrule
       \rowcolor{lightblue}
      \texttt{ILI}  & MSE & DLinear (2.169) & 1.750 & \textbf{-0.419} \\
      \rowcolor{lightblue}
      \texttt{ILI} & MAE & DLinear (1.041) & 0.816 & \textbf{-0.225} \\
        
      \bottomrule
    \end{tabular}
    \label{tab:absolute_diff_summary}
  \end{minipage}
  \hfill    
  % second minipage at 45%
  \begin{minipage}[t]{0.48\textwidth}
    \centering
    \begin{tabular}{lcccc}
      \toprule
      \textbf{Dataset} & \textbf{Pooling} & \textbf{Head} 
        & \textbf{MSE} & \textbf{MAE} \\
      \midrule
      \rowcolor{lightblue}
      \texttt{ETTh1} & none  & MLP    & \textbf{0.438} & \textbf{0.426} \\
      \texttt{ETTh1} & last time step  & MLP    & \underline{0.459} & \underline{0.443} \\
      \rowcolor{lightblue}
      \texttt{ETTh2} & none  & linear & \textbf{0.343} & \textbf{0.374} \\
      \texttt{ETTh2} & last time step  & linear & \underline{0.354} & \underline{0.386} \\
      \bottomrule
    \end{tabular}
    \label{tab:ablation_summary}
  \end{minipage}

  \caption{\texttt{CHARM}’s performance (forecasting) (left); probe ablation results (right)}
  \label{tab:combined_summary}
\end{table}
In \Cref{tab:ablation_summary} we summarize our ablation over different probing strategies categorized by "pooling" and "head" types (see \Cref{tab:combined_probe_ablations} for full ablation) to identify the most effective configuration of probing for the forecasting task. The results indicate that stacking embeddings across the entire lookback window consistently yields the best performance. Notably, we find that using only the embedding from the final time step performs competitively—with a modest increase in error ($\sim 5\%$ - \texttt{ETTh1}, $\sim 1\%$ - \texttt{ETTh2})—while offering substantial computational efficiency due to the reduced input size for the downstream probe.

\begin{figure*}[!h]
    \centering
    \includegraphics[width=\linewidth]{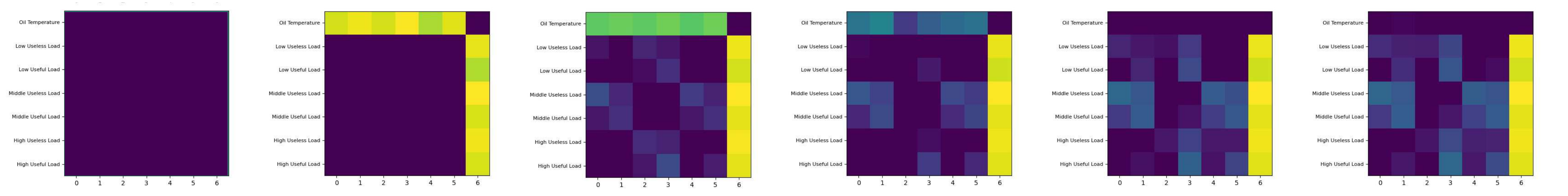}
    \caption{Evolution of Channel Gates for the \texttt{ETT} Dataset. A causal structure evolves over training, where the target causal variable \texttt{Oil Temperature} attends to all other independent channels but not vice versa. Extended discussion on evolution of channel gates can be found in \Cref{app:evolution_channel_gates}.}
    \label{fig:ett_evolve}
\end{figure*}

\section{Conclusion}
In this paper, we introduced \texttt{CHARM}, a novel foundation embedding model for multivariate time series. \texttt{CHARM} leverages a description-aware temporal convolutional network combined with a contextual attention mechanism that incorporates textual channel metadata, effectively capturing complex inter-channel interactions. Our architecture utilizes JEPA-inspired self-supervised learning, providing a robust alternative to traditional reconstruction-based or contrastive approaches and directly leveraging textual modality to enrich time series representations. Empirical evaluations across multiple benchmarks demonstrate \texttt{CHARM}’s effectiveness, setting new state-of-the-art performance in forecasting, classification, and anomaly detection tasks for representation learning methods. These findings underscore the practical utility and broad applicability of foundation embedding models for time series data. Importantly, \texttt{CHARM} also offers enhanced interpretability through heatmaps \Cref{fig:ett_evolve} that reveal cross-channel dynamics, making it not only performant but also insightful. Ablation studies (\Cref{app:ablation}) further validate the critical role of textual metadata, showing measurable gains in both downstream task performance and embedding quality, as quantified by effective rank.

As the first model to integrate granular textual information into foundational time series representation learning, \texttt{CHARM} opens several promising directions for future research. We anticipate deeper multimodal integrations and envision developing multi-task, multi-head architectures building upon \texttt{CHARM}’s foundational embeddings. Additionally, inspired by recent advancements in language modeling such as Retrieval-Augmented Generation (RAG), future work will focus on innovative interpretive frameworks that redefine how we interact with and interpret time series data.

\clearpage
\bibliography{iclr2025_conference}
\bibliographystyle{iclr2025_conference}

\appendix

\section{Related Work}

Historically, recurrent neural network architectures such as RNNs, LSTMs, and GRUs dominated time series modeling by capturing temporal dependencies through recursive hidden-state updates, achieving success across diverse tasks \citep{sal:20}. However, their sequential nature impeded parallelization, leading to slow training and difficulties in modeling long-range dependencies \citep{kal:16, pas:13, zho:21, kim:25}.

With the emergence of Transformer architectures \citep{vas:17}, significant advancements have occurred across multiple modalities, including images \citep{dosovitskiy2021imageworth16x16words}, audio \citep{gong2021astaudiospectrogramtransformer}, video \citep{arnab2021vivitvideovisiontransformer}, text \citep{devlin2018bert, radford2018improving}, and speech \citep{speech}. Inspired by these successes, the time series community has adopted Transformer-based approaches, leading to notable innovations tailored specifically for temporal data \citep{zhou2021informerefficienttransformerlong, zhou2022fedformerfrequencyenhanceddecomposed, ilbert2024samformerunlockingpotentialtransformers, wu:21, zhang2023crossformer}.

Simultaneously, self-supervised representation learning (SSRL), widely successful in domains such as vision and language, has demonstrated potential for extracting high-quality embeddings from vast amounts of unlabeled data. These embeddings facilitate downstream tasks—such as forecasting, classification, and anomaly detection—through lightweight task-specific heads. Analogous approaches have been adapted for time series, predominantly using contrastive self-supervised tasks \citep{yue:22, fra:24, fra:19, ton:21}. However, existing approaches typically produce models tailored to specific datasets, limiting their generalizability across arbitrary data sizes or channel configurations.

More recently, foundational models have revolutionized representation learning across natural language processing, computer vision, and audio \citep{dev:19, nus:25, ass:23, kir:23, bae:20, bro:20, rad:21}. In time series analysis, considerable progress has focused primarily on forecasting tasks \citep{das:24, woo:24, ans:24, liu:24}. Early foundational attempts predominantly addressed univariate series \citep{das:24, ans:24}, though recent advancements have successfully extended to multivariate settings with sophisticated cross-channel modeling techniques \citep{woo:24, liu:24}. Foundation embedding models specifically targeting time series representation learning have begun to emerge, leveraging reconstruction-based or next-step forecasting objectives \citep{goswami2024momentfamilyopentimeseries, gao2024unitsunifiedmultitasktime, trirat2024universaltimeseriesrepresentationlearning}. However, these approaches either focus on univariate series or treat multivariate data as independent channels, inadequately capturing complex inter-channel dynamics. This substantially limits the representational richness and effectiveness of these models in realistic scenarios. See \Cref{tab:comparison} for an overview of capabilities of key time series models.

\textbf{Joint-Embedding Predictive Architectures} have found notable success in visual domains by shifting the learning objective from pixel‐level reconstruction to latent‐space prediction. Extending this approach to video, Meta AI’s Video JEPA with Variance–Covariance Regularization (VJ‐VCR) \citep{drozdov2024video} predicts future frame embeddings in a learned representation space while enforcing variance and covariance constraints to prevent collapse; this model outperforms generative reconstruction baselines on downstream tasks such as action recognition and video retrieval by capturing high‐level spatiotemporal dynamics. Extensions such as MC‑JEPA \citep{bardes2023mcjepa} further demonstrate JEPA’s flexibility by jointly learning motion (optical flow) and content features within a shared encoder–predictor framework, matching or surpassing unsupervised optical flow benchmarks and improving downstream segmentation tasks. In multimodal settings, TI‑JEPA \citep{vo2025tijepa} integrates an energy‐based JEPA with cross‐modal encoders to align text and image embeddings, achieving superior results on multimodal sentiment analysis and visual question answering benchmarks by capturing complex semantic correspondences without reconstructing raw inputs. Complementing JEPA, bootstrapped embedding SSL methods like BYOL (“Bootstrap Your Own Latent”) \citep{grill2020bootstrap} train an online network to predict the target network’s representation of differently augmented views—updating the target via momentum averaging—and achieve strong results on ImageNet under linear evaluation without requiring negative pairs; this demonstrates that simple latent‐space prediction objectives can match or exceed contrastive and reconstruction‑based approaches in learning robust, generalizable representations. Together, these concrete instantiations highlight JEPA’s core advantage of filtering out low‑level noise and focusing learning on high‑level semantic structure, while bootstrapped SSL offers a practical, decoder‑free paradigm for self‑supervised representation learning., and motivate further exploration of these methods for time series.

\begin{table}[h]
\centering
\tiny
\begin{tabular}{|l|c|c|c|c|c|}
\hline
\textbf{Model} & \textbf{Multivariate} & \textbf{Channel Mixing} & \textbf{Equivariance} & \textbf{Foundational} & \textbf{Channel Aware} \\
\hline
Tloss & \cmark & \cmark & \xmark & \xmark & \xmark \\
TS2Vec & \cmark & \cmark & \xmark & \xmark & \xmark \\
TNC & \cmark & \cmark & \xmark & \xmark & \xmark \\
Autoformer & \cmark & \cmark & \xmark & \xmark & \xmark \\
FEDformer & \cmark & \cmark & \xmark & \xmark & \xmark \\
PatchTST & \cmark & \xmark & \xmark & \xmark & \xmark \\
CrossFormer & \cmark & \cmark & \xmark & \xmark & \xmark \\
iTransformer & \cmark & \cmark & \cmark & \cmark & \xmark \\
UniTS & \cmark & \cmark & \cmark & \cmark & \xmark \\
TimesFM & \xmark & -- & -- & \cmark & \xmark \\
MOIRAI & \cmark & \cmark & \cmark & \cmark & \xmark \\
MOMENT & \xmark & -- & \cmark & \cmark & \xmark \\
TREP & \cmark & \cmark & \xmark & \xmark & \xmark \\
\rowcolor{lightgray}
\texttt{CHARM} & \cmark & \cmark & \cmark & \cmark & \cmark \\
\hline
\end{tabular}
\captionsetup{justification=raggedright,singlelinecheck=false,position=bottom}
\caption{
\tiny
\textbf{a) Multivariate:} Can handle multivariate data\protect\footnotemark[1] \newline
\textbf{b) Channel Mixing:} Architecture enables learnable cross-channel interactions\protect\footnotemark[2] \newline
\textbf{c) Equivariance:} Permuting the channels by a perturbation $P$ ensures the outputs are also identically permuted. \newline
\textbf{d) Foundational:} Can flexibly accept data of any arbitrary number of channels or time window. \newline
\textbf{e) Channel Aware:} Uses sensor information to learn better representations.
}
\label{tab:comparison}
\end{table}
\footnotetext[1]{Multivariate here simply refers to whether a model can ingest multiple input channels, i.e. whether it can feasibly operate on a $T\times C$ data input, where $C>1$. This is independent of whether the model is able to learn channel interactions, which is explicitly outlined in the channel mixing column.}
\footnotetext[2]{We do not consider models that are fundamentally univariate, but perform late fusion of channels at the representation level (by pooling for example), to be capable of channel mixing.}

\section{Notation}
\label{app:notation}

We denote matrices and tensors using boldface capital letters (e.g., $\mathbf{T}$, $\mathbf{E}$), and adopt \emph{NumPy}-style indexing and slicing notation. Functions and operators are also denoted by bold capital letters, but are subscripted with $\theta$ to indicate parameterization, e.g., $\mathbf{E}_\theta$. The parameters $\theta$ may be learnable or fixed, depending on context. We reserve, $\mathbf W$ to represent the learnable weights in different layers of our architecture. An instance of a time series is represented as a tuple $\mathbf{t} = (\mathbf{T}, \mathbf{D}, \mathbf{pos})$. The first component, $\mathbf{T} \in \mathbb{R}^{T \times C}$, is a matrix of time series measurements, where $T$ denotes the number of time points and $C$ the number of channels. Each column $\mathbf{T}[:, i]$ corresponds to the uni-variate time series from channel $i$. The second component, $\mathbf{D}$, is an ordered list of length $C$, where each entry $\mathbf{D}[i]$ is a textual description of channel $i$, typically represented as a sentence or short passage. We assume that the descriptions in $\mathbf{D}$ are aligned with the corresponding columns of $\mathbf{T}$. The third component, $\mathbf{pos}$, represents the positional indices associated with the time series. We assume $\mathbf{pos} \in \mathbb{I}_+^T$ such that $|\mathbf{pos}| = T$. If not explicitly provided, we default to $\mathbf{pos} = [0, 1, \ldots, T{-}1]$. We denote the maximum time window size considered in our framework as $T_{\max}$.

\section{Implementation Details}
\label{app: impl_deets}

We attempt to follow the general set of best practices developed in the field of self-supervised learning, specifically those applicable to the Self-Distillation \citep{bal:23} family of algorithms. We outline the key details here;

\begin{enumerate}
    \item \textbf{Optimization Schedule} We use an AdamW optimizer to optimize our model. The learning rate follows a linear warmup followed by a cosine decay.
    
    \item \textbf{Weight Initialization} We use a fixed $\mathcal{N}(0,0.02)$ initialization which is commonly used in pretraining large transformer models \citep{olm:25}.

    \item \textbf{Weight Decay Scheduling} We use a cosine schedule for increasing the optimizer's weight decay over the course of training which has been shown to be crucial for training stability.
    
    \item \textbf{EMA Schedule for Target Encoder} We use an exponentially moving average with a momentum schedule that is increased gradually over the course of training.
    
\end{enumerate}

The weight decay scheduling and EMA schedule are identical to IJEPA \citep{ass:23}. Besides sweeping over a few learning rates, we perform no additional hyperparameter tuning on the rest of the hyperparameters due to limited compute, and list them in \Cref{tab:overall_hyperparams}.

\subsection{Rotary Position Embeddings}

Rotary Position Embeddings (RoPE)~\citep{su:24} differ from traditional additive positional embeddings in that they encode positional information by rotating the query and key vectors in a structured, position-dependent manner. Unlike fixed or learned additive embeddings, RoPE is applied at \textbf{each} layer of the self-attention computation, allowing the model to encode relative position information directly into the attention mechanism.

Let $\mathbf{Q}, \mathbf{K} \in \mathbb{R}^{B \times T \times D}$ denote the queries and keys, where $B$ is the batch size, $T$ is the sequence length, and $D$ is the hidden dimension. After linear projection and splitting into $H$ attention heads:

\[
\mathbf{Q}_h, \mathbf{K}_h \in \mathbb{R}^{B \times T \times d}, \quad \text{with } d = D / H
\]

RoPE applies a deterministic rotation to each head’s query and key vectors. For each position $t$ and dimension index $i$, the rotation is defined as:

\begin{align}
\text{RoPE}(\mathbf{x}_t)[2i] &= \mathbf{x}_t[2i] \cos(\theta_{t,i}) + \mathbf{x}_t[2i+1] \sin(\theta_{t,i}) \\
\text{RoPE}(\mathbf{x}_t)[2i+1] &= -\mathbf{x}_t[2i] \sin(\theta_{t,i}) + \mathbf{x}_t[2i+1] \cos(\theta_{t,i}) \\
\theta_{t,i} &= t \cdot \omega_i, \quad \omega_i = 10000^{-2i/d}
\end{align}

where $\mathbf{x}_t$ denotes the $t^{\text{th}}$ token's vector (query or key), and $\omega_i$ are predefined inverse frequency terms.

In our implementation, we operate on inputs of shape $\mathbf{X} \in \mathbb{R}^{B \times T \times C \times d}$, where $C$ represents the number of channels or sensors. To apply RoPE consistently across all channels, we broadcast the position encodings across the channel axis:

\begin{equation}
\widetilde{\mathbf{P}}_{b,t,c,:} = \mathbf{P}_{t,:}, \quad \forall\ b \in [1,B],\ c \in [1,C],\ t \in [1,T]
\end{equation}

or equivalently, using broadcasting semantics:

\begin{equation}
\widetilde{\mathbf{P}} = \mathbf{P}[t, :] \longrightarrow \mathbb{R}^{B \times T \times C \times d}
\end{equation}

This results in a broadcasted position encoding tensor $\widetilde{\mathbf{P}}$ where the same temporal position vector $\mathbf{P}_{t,:}$ is shared across all channels at time $t$, effectively associating the same position ID to multiple sensor tokens that occur at the same timestep.

\subsection{Text Convolution Layer}
\subsubsection{Contrast to other featurization methods}
Unlike typical TCNs, we concatenate activations across all intermediate layers to form a rich initial representation, see Figure~\ref{fig:tcn}. Our contextual TCN layer early in our model architecture closely relates to the concept of patching. Several recent foundation models for time series create non-overlapping static patches and project them through a single linear layer, e.g., \citep{das:24, nie:23, woo:24}. These approaches can be generalized by interpreting convolution kernels as learnable linear mappings applied to strided segments of the data. Thus, our TCN layer represents a generalized, channel-aware extension of the patching concept.

\subsubsection{Implementation}
To compute convolutions efficiently across all sensors and batches, we stack the convolutional kernels corresponding to each sensor description and reshape the input to treat the $B \times C \times H$ channels as independent time series. We then apply a grouped 1D convolution using \texttt{F.conv1d} with $B \times C \times H$ groups, where each element in the original $[B, T, C, H]$ input is treated as a separate time series along the time axis. This allows us to apply distinct filters for each batch, channel, and embedding dimension in parallel.

\subsubsection{Initialization}
Despite the effectiveness of this mechanism, careful numerical stabilization of the convolution kernels is essential. To achieve this, we first apply a non-parametric LayerNorm to $z$-normalize the sensor embeddings, $\mathbf{E}_d$. The projection matrix within the kernel network is then initialized using Xavier normal initialization \citep{glo:10}. Subsequently, we re-normalize the resulting kernels $\mathbf{W}_k$ as
$$
\mathbf{W}_k = \textbf{LayerNorm}(\mathbf{W}_k) \cdot \sqrt{\frac{2}{K}}
$$
Since our TCN layer employs GeLU nonlinearities, this initialization approach aligns with Kaiming initialization principles, \citep{he:15}, and ensures stable activations, preventing them from progressively exploding or vanishing across convolution layers.

\subsection{Model Sizing}
\label{app:model_size}
For the given hyperparameter set $N=8, d=128, \text{ff}_\text{dim}=4d$, our pretrained model is $\sim$7.1M parameters. 

\subsection{Additional Modifications to the transformer layers} \label{app:transformer}

In line with recent developments in large scale pretraining of transformer based architectures, we implement several modifications that diverge from the original transformer architecture. 

\paragraph{SwiGLU} We replace the regular feedforward layers with a SwiGLU feedforward layer.
\paragraph{QK-norm} We add a pre-attention layernorm to the queries and keys.
\paragraph{Rotary Position Embeddings} Instead of using sinusoidal positional embeddings, we use rotary positional embeddings which are applied on the queries and keys at every layer. The positional indices are provided through the $\textbf{pos}$ argument.
\paragraph{Reordering Sublayers} We experiment with using 3 approaches to assess the optimal configuration of the transformer sublayers.
\begin{equation}
\begin{cases}
    x=\text{norm}(x+\text{SubLayer}(x)) & \text{Post Norm}\\
    x=x+\text{SubLayer}(\text{norm}(x)) & \text{Pre Norm} \\
    x=x+\text{norm}(\text{SubLayer}(x)) & \text{Swin Norm} \\
\end{cases}
\end{equation}

In the case of Pre Norm and Swin Norm, we also experiment with adding LayerNorms in the main transformer branch every $n$ layers, to ensure further stability. 

\subsection{Efficient Computation Techniques}
\subsubsection{Slice And Tile Attention Layers}
\label{app: fast_attn}

To vectorize the process of generating the full $\mathbf{\bar{\Delta}}$ tensor, we provide the pytorch pseudocode versions of the naive and vectorized versions in \Cref{lst:attn_slow} and \Cref{lst:attn_fast}.
% Then, in the body of your paper:

\begin{figure}[ht]
\caption{Naïve attention‐weight matrix construction}
\label{lst:attn_slow}
\begin{lstlisting}[style=python]
def build_attention_weight_matrix(time_deltas: Tensor,
                                  T_proj: Tensor) -> Tensor:
    """
    Constructs the full attention weight matrix by explicit loops.
    Args:
        time_deltas: LongTensor, shape (T, T)
        T_proj:      Tensor, shape (B, C, C, 2*T - 1)
    Returns:
        attn:        Tensor, shape (B, C*T, C*T)
    """
    B, C, _, T1 = T_proj.shape
    T = time_deltas.size(0)
    assert 2 * T - 1 == T1

    attn = torch.zeros((B, C * T, C * T), device=T_proj.device)
    for i in range(T):
        for j in range(T):
            delta = time_deltas[i, j].item()
            block = T_proj[..., delta]    # (B, C, C)
            attn[..., i*C:(i+1)*C, j*C:(j+1)*C] = block
    return attn
\end{lstlisting}
\end{figure}

\begin{figure}[ht]
\caption{Fast attention‐weight matrix construction}
\label{lst:attn_fast}
\begin{lstlisting}[style=python]
def build_attention_weight_matrix_fast(time_deltas: Tensor,
                                       T_proj: Tensor) -> Tensor:
    """
    Block-wise assembly via tensor indexing and reshape.
    """
    B, C, _, T1 = T_proj.shape
    T = time_deltas.size(0)
    assert 2 * T - 1 == T1

    # 1) Flatten index grid
    flat_idx = time_deltas.view(-1)        # shape (T*T,)

    # 2) Gather all needed projection slices at once
    gathered = T_proj.index_select(dim=-1, index=flat_idx)
    #    result: (B, C, C, T*T)

    # 3) Reshape to (B, C, C, T, T)
    gathered = gathered.view(B, C, C, T, T)

    # 4) Reorder to (B, T, C, T, C)
    gathered = gathered.permute(0, 3, 1, 4, 2)

    # 5) Collapse blocks into (B, C*T, C*T)
    return gathered.contiguous().view(B, C * T, C * T)
\end{lstlisting}
\end{figure}

\begin{table}[!ht]
    \centering
    \begin{tabular}{c|c|c}
    \toprule
    \textbf{Category} & \textbf{Hyperparameter} & \textbf{Value} \\
    \midrule
    \multirow{9}{*}{\textbf{Optimization Schedule}} 
      & Optimizer   & AdamW \\
      & $\epsilon$   & 1e-8 \\
      & $\beta_1$ & 0.95 \\
      & $\beta_2$ & 0.99 \\
      & epochs & 100 \\
      & gradient clipping & 2.0 \\ 
      & $\lambda_1$, $\lambda_2$ & 1e-5 \\ 
      & batch size \footnote{See \Cref{app: dataloading} for additional discussions regarding the effective \texttt{batch size}.} & 4 \\
      & gradient accumulation & 2 \\
    \midrule
    \multirow{7}{*}{\textbf{Scheduler}} 
      & starting LR  & 1e-8 \\
      & final LR  & 1e-6 \\
      & starting weight decay  & 0.04 \\
      & final weight decay  & 0.4 \\
      & learning rate schedule  & linear warmup $\rightarrow$ cosine decay \\
      & weight decay schedule  & cosine \\
      & fraction of warmup epochs & 0.1 \\ 
      & scale factor \footnote{For better optimization dynamics, we use an "extended" schedule, similar to \citep{ass:23}, where the learning rate schedule is calculated for $\text{scale\_factor}\times T$ time steps, and then truncated to the original number of time steps $T$.} & 1.25 \\ 
    \midrule
    \multirow{4}{*}{\textbf{Data}} 
      & window size  & 512 \\
      & stride & 128 \\
      & minimum samples per dataset & 400 \\
      & maximum samples per dataset & 1000 \\

    \midrule
    \multirow{4}{*}{\textbf{SSL Task Parameters}} 
      & number of targets  & 4 \\
      & $C_{\min}$ & 0.3 \\
      & $C_{\max}$ & 0.4 \\
      & $T_{\min}$ & 0.1 \\
      & $T_{\max}$ & 0.2 \\
      
    \midrule
    \multirow{4}{*}{\textbf{JEPA Architecture}} 
      & encoder layers  & 8 \\
      & predictor layers  & 4 \\
      & encoder dim  & 128 \\
      & predictor dim  & 64 \\

    \midrule
    \multirow{5}{*}{\textbf{Model Architecture}} 
      & feedforward layer  & SwiGLU \\
      & ff\_dim\_multiplier  & 4 \\
      & attention dropout & 0.01\\
      & norm & non-parametric layernorm \\
      & attention configuration & pre-norm \\
      
    \bottomrule
    \end{tabular}
    \caption{Hyperparameters for full training pipeline}
    \label{tab:overall_hyperparams}
\end{table}

\section{JEPA}
\label{app: JEPA}

\subsection{Dataset Generation}
The core principle of JEPA-based self-superived training involves producing representations for two augmented views originating from the same data instance. JEPA training aims to minimize a discrepancy measure (e.g., $\ell_1$ or $\ell_2$) between these representations. In vision, these views commonly result from image augmentations such as jittering, masking, or cropping.

\begin{figure}
    \centering
    \includegraphics[width=0.7\linewidth]{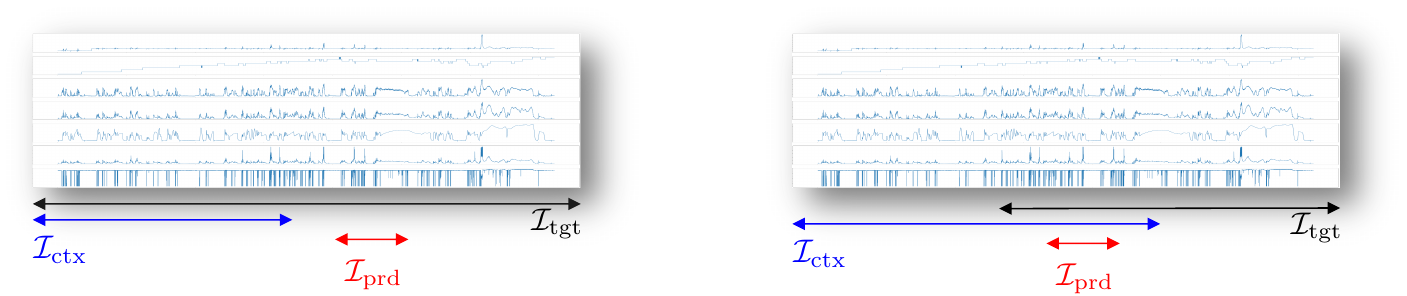}
    \caption{JEPA Tasks Visualized : Causal Prediction (left) Smoothing (right)}
    \label{fig:jepa_tasks}
\end{figure}

\Cref{fig:jepa_tasks} presents a visual representation of our JEPA tasks, which rely on learning 1) causal representations and 2) smoothing representations.

\subsection{JEPA Encoders -- Deep Dive}
In this section we dive a bit deeper into our implementation of the JEPA framework. We denote the TCN layer that featurizes our input time series as $\mathbf{F}$ and our encoder stack (of $N$ layers), as $\mathbf{E}$.

As outlined earlier, our featurizing layer converts a multivariate time series instance to an embedded version of the time series with the same leading dimensions, i.e.; 
$$\mathbf{F}:\mathbb{R}^{T\times C}\rightarrow \mathbb{R}^{T\times C\times H}$$

On the other hand our encoder ingests the embedded time series and returns a contextually embedded time series while maintaining the same output dimensions i.e.;

$$\mathbf{E}:\mathbb{R}^{T\times C\times H}\rightarrow \mathbb{R}^{T\times C\times H}$$

Given this notation, our 3 JEPA networks (Context, Target, Predictor) can be represented as:

\begin{align}
    \mathbf{Context}&\Rightarrow[\mathbf{F}\rightarrow \mathbf{E_1}] \\
    \mathbf{Target}&\Rightarrow[\mathbf{F}\rightarrow \mathbf{E_1}] \\
    \mathbf{Predictor}&\Rightarrow[\mathbf{DownProj}\rightarrow \mathbf{E_2}\rightarrow\mathbf{UpProj}]
\end{align}

Now, with this featurization and encoder layer stack, we provide a PyTorch style pseudocode of the JEPA framework, i.e. the data flow between the Context, Target, and Predictor encoders in \Cref{lst:ctx_tgt}, \Cref{lst:predictor}, and \Cref{lst:jepa}.

\begin{figure}[H]
\caption{Context and Target Network}
\label{lst:ctx_tgt}
\begin{lstlisting}[style=python]
class ContextTgtEncoder:
    def forward(self, x, ctx_idx):
        """
        x : [..., T, C] 
        """
        x = self.featurizer(x)
        for layer in self.encoder_layers:
            x = layer(x, ctx_idx)
        return x
\end{lstlisting}
\end{figure}

\begin{figure}[H]
\caption{Predictor Network}
\label{lst:predictor}
\begin{lstlisting}[style=python]
class Predictor:
    def forward(self, ctx_embeds, ctx_idx, target_idx):
        """
        embeds : [..., T, C, H]
        target_pos : [..., T2]
        """
        x = self.downproj(ctx_embeds)  # [..., T, C, H1]
        mask_tokens = broadcast(self.mask_token, target_idx)  # [..., T2, C, H1]
        
        x = concat([x, mask_tokens])  # [..., T+T2, C, H1]
        pos = concat([ctx_idx, target_idx])
        
        for layer in self.encoder_layers:
            x = layer(x, pos)
        x = self.upproj(x)  # [..., T+T2, C, H]
        x = x[..., -target_idx.size(-2):, :, :]  # [..., T2, C, H]
        return x
\end{lstlisting}
\end{figure}

\begin{figure}[!ht]
\caption{JEPA}
\label{lst:jepa}
\begin{lstlisting}[style=python]
class JEPA:
    def __init__(self):
        self.context_encoder = ContextTgtEncoder()
        self.target_encoder = copy_and_freeze_params(self.context_encoder)
        self.predictor = Predictor()
        
    def forward(self, x, ctx_idx, tgt_idx):
        """
        x : [..., T, C]
        """

        # get full embeddings
        full_embeds = self.target_encoder(x)

        # get context embeddings
        context_embeds = self.context_encoder(x[..., ctx_idx, :])

        # get predicted embeddings
        predicted_embeds = self.predictor(context_embeds, ctx_idx, tgt_idx)
        target_embeds = full_embeds[..., tgt_idx, :, :]

        # compute loss
        loss = loss_fn(predicted_embeds, target_embeds)
        return loss
\end{lstlisting}
\end{figure}

\section{Hardware} \label{hardware}
We use a cluster of 8 80GB NVIDIA A100 GPUs. We use Distributed Data Parallelism to speed up training, along with bf-16 mixed precision. Our models are implemented in pytorch \citep{paszke2019pytorchimperativestylehighperformance}, and training is done with pytorch lightning \citep{fal_torch}. We handle our configuration management using gin configs. 

\section{Limitations} 
The key mechanism in our architecture enables learning channel interaction dynamics using sensor descriptions and real time series data. A major bottleneck for our model's efficacy is the access to high quality, descriptive descriptions. Further, computationally our model uses a self attention mechanism which is $O(C^2T^2)$, which can be prohibitive for large datasets with multiple channels. Additionally, our model does not confer direct advantages over univariate modeling without informative covariates.

\section{Datasets} \label{app:datasets}
Here we provide a list of dataset sources we used to train our model. Wherever sensor names were not readily available, we manually curate the sensor descriptions from the dataset specifications.
\paragraph{UEA Dataset}
    The UEA Dataset is a popular publicly available dataset used for benchmarking time series classification algorithms. We restrict ourselves to a subset of the full 30 datasets, as not all of them have meaningful sensor descriptions. For a few of the datasets within UEA, we manually annotate the descriptions based on the official paper \citep{uea}.
    \paragraph{Liu-ICE Machine Fault Dataset}
    The Liu-ICE Machine Fault Dataset is a real world fault diagnosis dataset which consists of data collected from an internal combustion engine test bench. The dataset consists of multiple different kinds of fault scenarios, and comes with a publicly available benchmark.
    \paragraph{Electricity Transformer Dataset}
    The Electricity Transformer Dataset (ETTDataset/ETDataset) is a widely used dataset for time series forecasting, which contains data of dynamic power loads in an electric power grid located in China. This dataset contains 4 sub-datasets (\texttt{ETTh1}, \texttt{ETTh2}, \texttt{ETTm1}, \texttt{ETTm2}), which operate at different granularities.
    \paragraph{Weather}
    The Weather Dataset from the MPI is a real world dataset of meteorological indicators for the year of 2020.
    \paragraph{Electricity}
    The Electricity dataset contains hourly consumption from multiple consumers from 2012 to 2014.
    \paragraph{Illness}
    The Illness Dataset includes weekly records for patients suffering from influenza like illnesses collected by the CDC.
    \paragraph{SKAB - Skoltech Anomaly Benchmark Dataset}
    The SKAB dataset is designed for evaluating anomaly detection, targeted at two main problems : outlier detection and changepoint detection
    \paragraph{Gas Sensor Array Modulation}
    The Gas Sensor Array Modulation from the UCI Machine Learning Repository is collection of time‐series recordings obtained from an array of metal‐oxide gas sensors.
    \paragraph{Machinery Fault Dataset}
    The Machinery Fault Dataset comprises six different simulated states: normal function, imbalance fault, horizontal and vertical misalignment faults and, inner and outer bearing faults from a machinery fault simulator.
    \paragraph{Metro PT-3 Dataset}
    The MetroPT-3 dataset is a multivariate time series collection created for predictive maintenance in the railway industry. It consists of over 1.5 million records (instances) captured at 1Hz from a train compressor’s Air Production Unit (APU) over the period from February to August 2020.
    \paragraph{Unleashing the Power of Wearables}
    The Human Activity Recognition Trondheim (HEART) dataset is a professionally annotated collection designed for developing machine learning algorithms capable of recognizing human activities in a free-living environment. Created at the Norwegian University of Science and Technology (NTNU), it features 22 subjects who wore two 3-axis Axivity AX3 accelerometers for approximately 2 hours each while performing various daily tasks. The sensors were placed on the right thigh and lower back, providing multivariate time series data sampled at 50Hz.
    \paragraph{Predictive Maintenance of Hydraulic Systems}
    The Predictive Maintenance of Hydraulic Systems dataset contains multivariate time series data collected from a hydraulic test rig. This dataset includes sensor readings—such as pressures, volume flows, temperatures, and more—recorded during load cycles of the hydraulic system.

We provide a summary of the specifications of each dataset in \Cref{tab:datasets_categorized}. If a dataset is present in a downstream benchmark, we only include the defined "train" subset of the full dataset, to prevent the model from optimizing an SSL loss over the test dataset samples.

\begin{table}[ht]
\centering

\begin{tabular}{lrr}
\toprule
\textbf{Dataset Name} & \textbf{\#Timestamps} & \textbf{\#Channels} \\
\midrule
\multicolumn{3}{l}{\textbf{Open-Source/Kaggle Datasets}} \\
Appliances Energy Prediction                & 19,735       & 26  \\
Gas Sensor Array Temperature Modulation    & 3,843,160    & 19  \\
Household Electric Power Consumption       & 2,075,259    & 7   \\
Machinery Fault Diagnosis                   & 487,748,049  & 8   \\
MetroPT-3 Dataset                           & 1,516,948    & 15  \\
Predictive Maintenance of Hydraulics System & 132,300      & 17  \\
SKAB - Skoltech Anomaly Benchmark           & 46,860       & 8   \\
Unleashing the Power of Wearables           & 6,461,328    & 6   \\
Liu                                         & 288,623      & 10  \\

\midrule
\multicolumn{3}{l}{\textbf{UEA Datasets} \footnotemark} \\
NATOPS                                      & 9180          & 24  \\
Epilepsy                                    & 28222         & 3   \\
Articulary Word Recognition                 & 39600          & 9   \\
UWave Gesture Library                       & 37800          & 3   \\
Cricket                                     & 129276          & 6   \\
ERing                                       & 1950           & 4   \\
Character Trajectories                      & 169218        & 3   \\
Finger Movements                            & 15800          & 28  \\
SelfRegulation SCP1                         & 240128          & 6   \\
Basic Motions                               & 4000           & 6   \\
Atrial Fibrillation                         & 9600           & 2   \\
Hand Movement Direction                     & 64000          & 10  \\
Handwriting                                 & 22800          & 3   \\
Libras                                      & 8100          & 2   \\
LSST                                        & 88,524        & 6   \\
Racket Sports                               & 4530          & 6   \\

\midrule
\multicolumn{3}{l}{\textbf{Forecasting Benchmark Datasets}} \\
\texttt{ETTh1}                                       & 17,420       & 7   \\
\texttt{ETTh2}                                       & 17,420       & 7   \\
\texttt{ETTm1}                                       & 69,680       & 7   \\
\texttt{ETTm2}                                       & 69,680       & 7   \\
Weather                                     & 52,696       & 21  \\
Illness                                     & 966          & 7   \\

\bottomrule
\end{tabular}
\caption{Overview of datasets categorized into Open-Source/Kaggle, UEA, and Forecasting benchmark datasets.}
\label{tab:datasets_categorized}
\end{table}

\footnotetext{The UEA Datasets are provided as windowed instances, i.e. they are not hosted as contiguous, chronological blocks of shape $T\times C$, but rather stored as $N\times T'\times C$. Here, we compute the "\# of timesteps" as $N \times T'$, although there may be redundant overlaps based on how the data was collected and labelled.}

\section{Data Loading} \label{app: dataloading}
To enable efficient data loading, we perform under/over sampling to balance the datasets. The degree of under/over sampling is controlled by the $t_1$ : \texttt{min\_samples\_per\_dataset} and $t_2$ : \texttt{max\_samples\_per\_dataset} parameters, which upsamples or downsamples the data if the number of samples is either $<t_1$ or $>t_2$ respectively. 

Following this, each dataset is handled by its own dataloader, which cyclically yields batches of data from each dataset at every training step. This is handled internally by \texttt{pytorch lightning}'s \texttt{CombinedLoader} method, which yields a batch from each dataloader (if the iterator is not yet exhausted). As a result, our \texttt{effective batch size} \footnotemark per step is now computed as :
\begin{eqnarray}
(\texttt{batch size}) \times (\text{\# of GPUs}) \times (\text{\# of datasets}) \times (\texttt{grad\_accum\_steps})
\end{eqnarray}

\footnotetext{The "\# of datasets" technically refers to the number of unexhausted datasets on that training step, as each dataset has a different number of samples.}
At the beginning of every epoch, we reload all datasets, which results in fresh under/over sampling indices. This enables the support of larger datasets to be incrementally covered over multiple epochs of training.

The JEPA tasks are randomly sampled after the datasets are sampled, which results in fresh context and target masks for repeated samples. This avoids the exact same sample being repeated several times in an epoch for underrepresented datasets, due to stochasticity in how the masks are generated.

\section{Experiments}
\label{app: experiments}

\subsection{Multivariate Classification}
\label{app: classification}

We evaluate our model on the popular UEA Dataset which serves as a standard time series classification benchmark for multivariate data. We consider a subset of 15 UEA datasets that cover a diverse set of tasks and domains.

Given a labelled time series data instance $(X,y)$, where $X$ has is a multivariate time series, and $y$ corresponds to a supervised label corresponding to $X$, our goal is to minimize test error, i.e. $\epsilon=\mathbb{E}_{(x,y)\sim\mathbb{P}_{(x,y)}}[1(h(x)\neq y)]$. For a downstream classification task, we use our frozen pre-trained encoder to extract embeddings $Z\in\mathbb{R}^{T\times C\times H}$, where $T$ and $C$ refer to the number of timestamps and channels in the data instance. The embeddings $Z$ are then mean pooled, to return a final $\bar{Z}\in \mathbb{R}^H$, which is used as an input to a standard off-the-shelf classification algorithm. To ensure a fair comparison with other representation learning models, we use a Support Vector Machine, similar to \citep{fra:24}, and perform a grid search over the SVM's hyperparameters \Cref{tab: svm_grid} for each dataset. We tune the hyperparameters for each UEA dataset separately. Additionally, we probe the model's checkpoints as it evolves during training, and report the best performing scores in \Cref{tab:tsc_performance}.

Note, that unlike other representation learning models for time series which are dataset specific, our model is pretrained on a global dataset, and then used to generate representations for each dataset's evaluations. 

\begin{table}[!ht]
    \centering
    \begin{tabular}{c|c}
    \toprule
    \textbf{Hyperparameter} & \textbf{Value} \\
    \midrule
    \texttt{C} & \{0.0001, 0.001, 0.01, 0.1, 1, 10, 100, 1000, 10000\} \\
    \texttt{kernel} & \{'rbf'\} \\
    \texttt{degree} & \{3\} \\
    \texttt{gamma} & \{'scale'\} \\
    \texttt{coef0} & \{0\} \\
    \texttt{shrinking} & \{True\} \\
    \texttt{probability} & \{False\} \\
    \texttt{tol} & \{0.001\} \\
    \texttt{cache\_size} & \{200\} \\
    \texttt{class\_weight} & \{None\} \\
    \texttt{verbose} & \{False\} \\
    \texttt{max\_iter} & \{10000000\} \\
    \texttt{decision\_function\_shape} & \{'ovr'\} \\
    \texttt{random\_state} & \{None\} \\
    \bottomrule
    \end{tabular}
    \caption{SVM Hyperparameter Grid}
    \label{tab: svm_grid}
\end{table}

\begin{table}[H]
\centering
\resizebox{\textwidth}{!}{
\begin{tabular}{lccccccc}
\hline
\textbf{Dataset} & \textbf{DTW} & \textbf{TS2Vec} & \textbf{T‑Loss} & \textbf{TNC} & \textbf{TS‑TCC} & \textbf{T‑Rep} & \textbf{CHARM} \\
\hline
AtrialFibrillation            & 20.0  & 29.0  & 13.0  & 13.0  & 27.0  & 35.0  & \textbf{47.0} \\
Articulary/WordRecognition    & \textbf{98.7}  & 97.4  & 94.3  & 97.3  & 95.3  & 96.8  & 98.6 \\
BasicMotions                  & \textbf{100.0} & \textbf{100.0} & \textbf{100.0} & \textbf{100.0} & \textbf{100.0} & \textbf{100.0} & 97.5 \\
CharacterTrajectories         & 98.9  & 98.8  & \textbf{99.3}  & 96.7  & 98.5  & 98.9  & 98.1 \\
Cricket                       & \textbf{100.0} & 95.3  & 97.2  & 95.8  & 91.7  & 95.8  & 95.8 \\
ERing                         & 13.3  & 89.9  & 13.3  & 85.2  & 90.4  & 94.3  & \textbf{96.2} \\
Epilepsy                      & 96.4  & 96.3  & 97.1  & 95.7  & 95.7  & 97.0  & \textbf{97.8} \\
FingerMovements               & 53.0  & 49.5  & 58.0  & 47.0  & 46.0  & 49.5  & \textbf{59.0} \\
HandMovementDirection         & 23.1  & 41.8  & 35.1  & 32.4  & 24.3  & 53.6  & \textbf{54.0} \\
Handwriting                   & 28.6  & 46.3  & 45.1  & 24.9  & \textbf{49.8}  & 41.4  & 33.0 \\
LSST                          & 55.1  & 55.8  & 50.9  & 59.5  & 47.4  & 52.6  & \textbf{59.8} \\
Libras                        & 87.0  & 85.9  & \textbf{88.3}  & 81.7  & 82.2  & 82.9  & 83.0 \\
NATOPS                        & 88.3  & 89.7  & \textbf{91.7}  & 91.1  & 82.2  & 80.4  & 82.2 \\
RacketSports                  & 80.2  & \textbf{89.3}  & 85.5  & 77.5  & 81.6  & 88.3  & 86.0 \\
SelfRegulationSCP1            & 77.5  & 79.0  & \textbf{84.3}  & 79.9  & 82.3  & 81.9  & 82.0 \\
UWaveGestureLibrary           & 90.3  & 87.5  & 87.5  & 75.9  & 75.3  & 88.5  & \textbf{91.2} \\
\hline
Avg Accuracy                  & 0.694  & 0.769  & 0.712  & 0.721  & 0.731  & 0.773  & \textbf{0.788} \\
Ranks 1st                     & 3       & 2       & 5       & 1       & 2       & 1       & \textbf{7}     \\
\hline
\end{tabular}
}
\caption{
Performance on UEA Multivariate Classification Datasets \\
If more than 1 model has the same score on a particular dataset, we consider both models to rank 1st.}
\label{tab:tsc_performance}
\end{table}

\subsection{Anomaly Detection}
\label{app:anomaly_detection}
To evaluate our model's embeddings on anomaly detection benchmarks, we use the popular Skoltech Anomaly Benchmark suite \citep{skab}, henceforth referred to as SKAB, which consists of a pointwise anomaly detection task using data from 8 sensors attached to a mechanical testbed. The anomaly detection task consists of inputs of the form $(X,y)$, where $X\in \mathbb{R}^{T\times C}$ and $y\in \mathbb{R}^T$, where each $y_i\in \{0,1\}$ where an annotation of $1$ indicates an anomaly. 

The SKAB suite by default uses reconstruction based methods to detect anomalies. To mimic the reconstruction based methods, we add a single linear layer on top of our embedding outputs to reconstruct the real data. I.e., given an input $X\in \mathbb{R}^{T\times C}$, we first embed the data point-wise, $Z\in\mathbb{R}^{T\times C \times H}$, and then apply a linear layer $W\in\mathbb{R}^{H\times1}$, along with a learnable bias $b$ for each channel, $c\in C$. This results in a reconstruction of the input, denoted by $\hat{X}\in\mathbb{R}^{T\times C}$. As the embedding model is not trained for a reconstruction task, it is infeasible for the model to collapse to learn an identity mapping, which would make the reconstruction objective trivial (from a training point of view) and consequently overfit to the train set. 

Note, the embedding model is \textbf{frozen} during this evaluation protocol, and the learnable parameters are trained using backpropagation to minimize a reconstruction loss. We do not tune any hyperparameters for training this probe, and mention our selected hyperparameters in \Cref{tab:anomaly_hyperparams}.

\begin{table}[H]
    \centering
    \begin{tabular}{c|c}
    \toprule
    \textbf{Hyperparameter} & \textbf{Value} \\
    \midrule
      Optimizer   & AdamW \\
      Weight Decay   & None \\
      Learning Rate & 1e-3 \\
      Epochs & 1000 \\
    \bottomrule
    \end{tabular}
    \caption{Hyperparameters to train reconstruction head for anomaly detection}
    \label{tab:anomaly_hyperparams}
\end{table}

We compare our model's performance to other reconstruction based models tested with the same evaluation protocol on the SKAB test suite. This involves splitting the data into several train and test sets, where each instance in the train and test set is trained with a fresh model to reconstruct the training dataset, i.e. minimize MSE on the reconstruction task $||x_{\text{train}}-\hat{x}_{\text{train}}||_2$, and then evaluated by computing the reconstruction of the corresponding test instance, $\hat{x}_{\text{test}}$. Based on the train set reconstruction, we compute the Upper Control Limit (UCL), based on the 99th percentile quantiles, and apply an adjustment factor of $\frac{4}{3}$. Then, for the reconstructed test data $\hat{x}_{\text{test}}$, we classify anomalies if the absolute values of the residuals, i.e. $||x_{\text{test}}-\hat{x}_{\text{test}}||$ lie outside the UCL limit.

\subsection{Forecasting}
\label{app:forecasting}
Forecasting tasks consist of taking a window of time series data and predicting future time steps. Formally, given an input of dimensions $(T_h\times C)$, where $T_h$ denotes the "lookback" horizon, the goal is to predict the future $T_f$ time steps. In our evaluation setup, we use the embeddings of the input horizon data, and extract 1a) all embeddings for each channel, 2a) last time step's embedding for each channel, and 3a) mean pooled embeddings for each channel. Following this, we add a single 1b) linear layer, 2b) 2 layer MLP with ReLU non-linearities, and train the model to minimize an aggregate loss metric\footnote{$\mathbf{loss}=\frac{\mathbf{MSE}+\mathbf{MAE}}{2}$} between the predicted and true values for each channel. We report the mean and standard deviation of the best performing scores over 5 runs in \Cref{tab:ETT_results}.

In line with our frozen evaluation protocols, we keep the embedding model frozen, and train the linear probe using backpropagation. We outline the probe hyperparameters in \Cref{tab:forecast_hyperparams}. Further, we report the results of our probing ablations for \texttt{ETTh1} and \texttt{ETTm2} in \Cref{tab:combined_probe_ablations}.

The train/valid/test split is identical to the standard protocol in the other baselines we compare with, which is a 6/2/2 split for the ETT datasets, and a 7/1/2 split for all other datasets.

To ensure a fair comparison, we use the standard set of lookback horizon, and future horizon values for all forecasting datasets, as outlined in \Cref{tab:ETT_specs}.

\begin{table}[H]
    \centering
    \begin{tabular}{c|c|c}
    \toprule
      \textbf{Dataset}   &  \textbf{Lookback Horizon} $T_h$ & \textbf{Target Horizon} $T_f$\\
    \midrule
      \texttt{ETTh1}  & 96 & {24, 48, 168, 336, 720} \\
      \texttt{ETTh2}  & 96 & {24, 48, 168, 336, 720} \\
      \texttt{ETTm1}  & 96 & {24, 48, 96, 288, 672} \\
      \texttt{ETTm2}  & 96 & {24, 48, 96, 288, 672} \\
    \midrule
      Weather & 96 & 96, 192, 336, 672 \\
      Exchange Rate & 96 & 96, 192, 336, 672 \\
      Illness & 96 & 24, 36, 48, 60 \\
    \bottomrule
    \end{tabular}
    \caption{Forecasting Task Specifications}
    \label{tab:ETT_specs}
\end{table}

\begin{table}[H]
    \centering
    \begin{tabular}{c|c}
    \toprule
    \textbf{Hyperparameter} & \textbf{Value} \\
    \midrule
      Optimizer   & AdamW \\
      Weight Decay   & [1e-2, 1e-4] \\
      Learning Rate & [1e-2, 1e-4] \\
      Epochs & 1000 \\
      LR Schedule & ReduceLROnPlateau \\
      Reduction Factor & 0.1 \\
      Early Stopping : Patience & 50 \\
      Early Stopping : Tolerance & 1e-6 \\
      
    \bottomrule
    \end{tabular}
    \caption{Hyperparameters to train prediction head for forecasting tasks}
    \label{tab:forecast_hyperparams}
\end{table}

\subsubsection{ETT}

\begin{table}[H]
\centering
\resizebox{\textwidth}{!}{%
\begin{tabular}{llcccccccccccc}
\toprule
\textbf{Dataset} & \textbf{H}
  & \multicolumn{2}{c}{\textbf{T-Rep}}
  & \multicolumn{2}{c}{\textbf{TS2Vec}}
  & \multicolumn{2}{c}{\textbf{Informer}}
  & \multicolumn{2}{c}{\textbf{TCN}}
  & \multicolumn{2}{c}{\textbf{Linear}}
  & \multicolumn{2}{c}{\textbf{CHARM}} \\
\cmidrule(lr){3-4} \cmidrule(lr){5-6} \cmidrule(lr){7-8} \cmidrule(lr){9-10} \cmidrule(lr){11-12} \cmidrule(lr){13-14}
&
& \textbf{MSE} & \textbf{MAE}
& \textbf{MSE} & \textbf{MAE}
& \textbf{MSE} & \textbf{MAE}
& \textbf{MSE} & \textbf{MAE}
& \textbf{MSE} & \textbf{MAE}
& \textbf{MSE} & \textbf{MAE} \\
\midrule
\multirow{5}{*}{\textbf{\texttt{ETTh1}}}
  & 24  & \underline{0.511} & \underline{0.496}  & 0.575 & 0.529  & 0.577 & 0.549  & 0.767 & 0.612  & 0.873 & 0.664  & 
  \textbf{0.310}\textsubscript{$\pm0.0013$} 
  & \textbf{0.350}\textsubscript{$\pm0.0008$} \\
  
  & 48  & \underline{0.546} & \underline{0.524}  & 0.608 & 0.553  & 0.685 & 0.625  & 0.713 & 0.617  & 0.912 & 0.689  
  & \textbf{0.358}\textsubscript{$\pm0.0017$}  
  & \textbf{0.376}\textsubscript{$\pm0.0009$}  \\
  
  & 168 & \underline{0.759} & \underline{0.649}  & 0.782 & 0.659  & 0.931 & 0.752  & 0.995 & 0.738  & 0.993 & 0.749  
  & \textbf{0.451}\textsubscript{$\pm0.0008$} 
  & \textbf{0.430}\textsubscript{$\pm0.0016$} \\
  
  & 336 & \underline{0.936} & \underline{0.742}  & 0.956 & 0.753  & 1.128 & 0.873  & 1.175 & 0.800  & 1.085 & 0.804  
  & \textbf{0.517}\textsubscript{$\pm0.0024$} 
  & \textbf{0.466}\textsubscript{$\pm0.0013$} \\
  
  & 720 & \underline{1.061} & \underline{0.813}  & 1.092 & 0.831  & 1.215 & 0.896  & 1.453 & 1.311  & 1.172 & 0.863  
  & \textbf{0.546}\textsubscript{$\pm0.0036$} 
  & \textbf{0.498}\textsubscript{$\pm0.0028$} \\
  
\cmidrule(lr){1-14}
  & Avg   & 0.763 & 0.645  & 0.803 & 0.665  & 0.907 & 0.739  & 1.021 & 0.816  & 1.007 & 0.754  & \textbf{0.436} & \textbf{0.424} \\
  & Wins  & 0     & 0      & 0     & 0      & 0     & 0      & 0     & 0      & 0     & 0      & \textbf{5}     & \textbf{5}     \\
\midrule
\multirow{5}{*}{\textbf{\texttt{ETTh2}}}
  & 24  & 0.560 & 0.565  & \underline{0.448} & 0.506  & 0.720 & 0.665  & 1.365 & 0.888  & 0.463 & \underline{0.498}  
  & \textbf{0.186}\textsubscript{$\pm0.0005$} 
  & \textbf{0.267}\textsubscript{$\pm0.0002$} \\
  
  & 48  & 0.847 & 0.711  & 0.685 & 0.642  & 1.457 & 1.001  & 1.395 & 0.960  & \underline{0.614} & \underline{0.588}  
  & \textbf{0.242}\textsubscript{$\pm0.0014$} 
  & \textbf{0.303}\textsubscript{$\pm0.0006$} \\
  
  & 168 & 2.327 & 1.206  & 2.227 & 1.164  & 3.489 & 1.515  & 3.166 & 1.407  & \underline{1.738} & \underline{1.016}  
  & \textbf{0.391}\textsubscript{$\pm0.0015$} 
  & \textbf{0.396}\textsubscript{$\pm0.0004$} \\
  
  & 336 & 2.665 & 1.324  & 2.803 & 1.360  & 2.723 & 1.340  & 3.256 & 1.481  & \underline{2.198} & \underline{1.173}  
  & \textbf{0.430}\textsubscript{$\pm0.0024$} 
  & \textbf{0.427}\textsubscript{$\pm0.0011$} \\
  
  & 720 & 2.690 & 1.365  & 2.849 & 1.436  & 3.467 & 1.473  & 3.690 & 1.588  & \underline{2.454} & \underline{1.290}  
  & \textbf{0.470}\textsubscript{$\pm0.0019$} 
  & \textbf{0.466}\textsubscript{$\pm0.0007$} \\

\cmidrule(lr){1-14}
  & Avg   & 1.818 & 1.034  & 1.802 & 1.022  & 2.371 & 1.199  & 2.574 & 1.265  & \underline{1.493} & \underline{0.913}  & \textbf{0.344} & \textbf{0.372} \\
  & Wins  & 0     & 0      & 0     & 0      & 0     & 0      & 0     & 0      & 0     & 0      & \textbf{5}     & \textbf{5}     \\
\midrule
\multirow{5}{*}{\textbf{\texttt{ETTm1}}}
  & 24  & 0.417 & 0.420  & 0.438 & 0.435  & \underline{0.323} & \underline{0.369}  & 0.324 & 0.374  & 0.590 & 0.505  & \textbf{0.218\textsubscript{$\pm0.0006$}} & \textbf{0.283\textsubscript{$\pm0.0008$}} \\
  
  & 48  & 0.526 & 0.484  & 0.582 & 0.553  & 0.494 & 0.505  & \underline{0.477} & \underline{0.450}  & 0.813 & 0.637  & \textbf{0.282\textsubscript{$\pm0.0004$}} & \textbf{0.324\textsubscript{$\pm0.0003$}} \\
  
  & 96  & \underline{0.573} & \underline{0.516}  & 0.602 & 0.537  & 0.678 & 0.614  & 0.636 & 0.602  & 0.866 & 0.654  & \textbf{0.316\textsubscript{$\pm0.0001$}} & \textbf{0.347\textsubscript{$\pm0.0009$}} \\
  
  & 288 & \underline{0.648} & \underline{0.577}  & 0.709 & 0.610  & 1.056 & 0.786  & 1.270 & 1.351  & 0.929 & 0.697  & \textbf{0.395\textsubscript{$\pm0.0001$}} & \textbf{0.391\textsubscript{$\pm0.0015$}} \\
  
  & 672 & \underline{0.758} & \underline{0.649}  & 0.826 & 0.687  & 1.192 & 0.926  & 1.381 & 1.467  & 1.008 & 0.746  & \textbf{0.482\textsubscript{$\pm0.0045$}} & \textbf{0.441\textsubscript{$\pm0.0039$}} \\
  
\cmidrule(lr){1-14}
  & Avg   & \underline{0.584} & \underline{0.529}  & 0.631 & 0.564  & 0.749 & 0.640  & 0.818 & 0.849  & 0.841 & 0.648  & \textbf{0.338} & \textbf{0.357} \\
  & Wins  & 0     & 0      & 0     & 0      & 0     & 0      & 0     & 0      & 0     & 0      & \textbf{5}     & \textbf{5}     \\
\midrule
\multirow{5}{*}{\textbf{\texttt{ETTm2}}}
  & 24  & 0.172 & 0.293  & 0.189 & 0.310  & \underline{0.147} & \underline{0.277}  & 1.452 & 1.938  & 0.275 & 0.364  & \textbf{0.099\textsubscript{$\pm0$}} & \textbf{0.192\textsubscript{$\pm0.0001$}} \\
  & 48  & 0.263 & 0.377  & \underline{0.256} & \underline{0.369}  & 0.267 & 0.389  & 2.181 & 0.839  & 0.363 & 0.434  & \textbf{0.131\textsubscript{$\pm0.0001$}} & \textbf{0.223\textsubscript{$\pm0.0002$}} \\
  & 96  & 0.397 & 0.470  & 0.402 & 0.471  & \underline{0.317} & \underline{0.411}  & 3.921 & 1.714  & 0.441 & 0.484  & \textbf{0.172\textsubscript{$\pm0.0003$}} & \textbf{0.253\textsubscript{$\pm0.0004$}} \\
  & 288 & 0.897 & 0.733  & 0.879 & 0.724  & 1.147 & 0.834  & 3.649 & 3.245  & \underline{0.754} & \underline{0.664}  & \textbf{0.284\textsubscript{$\pm0.0001$}} & \textbf{0.326\textsubscript{$\pm0.0005$}} \\
  & 672 & 2.185 & 1.144  & 2.193 & 1.159  & 3.989 & 1.598  & 6.973 & 1.719  & \underline{1.796} & \underline{1.027}  & \textbf{0.403\textsubscript{$\pm0.0042$}} & \textbf{0.400\textsubscript{$\pm0.0022$}} \\
\cmidrule(lr){1-14}
  & Avg   & 0.783 & 0.603  & 0.784 & 0.607  & 1.173 & 0.702  & 3.635 & 1.891  & \underline{0.726} & \underline{0.595}  & \textbf{0.217} & \textbf{0.278} \\
  & Wins  & 0     & 0      & 0     & 0      & 0     & 0      & 0     & 0      & 0     & 0      & \textbf{5}     & \textbf{5}     \\
\midrule
  & \textbf{Overall} & \underline{0.987} & \underline{0.703}  & 1.005 & 0.714  & 1.300 & 0.820  & 2.012 & 1.205  & 1.017 & 0.727  & \textbf{0.333} & \textbf{0.358} \\
\bottomrule
\end{tabular}%
}
\caption{
\\
Long horizon forecasting results on ETT datasets across different horizons. Input length = $96$.
\\Lower is better. \textbf{Bold} = best, \underline{Underline} = second best.}
\label{tab:ETT_results}
\end{table}

\begin{table}[ht]
\centering
\resizebox{\textwidth}{!}{\begin{tabular}{lll*{5}{cc}}
\toprule
\textbf{Dataset} & \textbf{Pool} & \textbf{Head} 
& \multicolumn{2}{c}{\textbf{24}} 
& \multicolumn{2}{c}{\textbf{48}} 
& \multicolumn{2}{c}{\textbf{168}} 
& \multicolumn{2}{c}{\textbf{336}} 
& \multicolumn{2}{c}{\textbf{720}} \\
\cmidrule(lr){4-5}
\cmidrule(lr){6-7}
\cmidrule(lr){8-9}
\cmidrule(lr){10-11}
\cmidrule(lr){12-13}
& & & MSE & MAE & MSE & MAE & MSE & MAE & MSE & MAE & MSE & MAE \\
\midrule
\multirow{6}{*}{\texttt{ETTh1}} 
  & \multirow{2}{*}{\textbf{none}$^{1a}$} 
    & linear$^{1b}$ & \textbf{0.31} & \textbf{0.35} & 0.36 & \textbf{0.38} & 0.45 & \textbf{0.43} & 0.52 & 0.47 & 0.55 & \textbf{0.50} \\
  & & MLP$^{2b}$    & 0.32 & 0.36 & \textbf{0.37} & 0.38 & \textbf{0.46} & 0.44 & \textbf{0.50} & \textbf{0.46} & \textbf{0.54} & \textbf{0.50} \\

  \cmidrule(lr){2-13}
  
  & \multirow{2}{*}{\textbf{last time step}$^{2a}$} 
    & linear$^{1b}$ & 0.39 & 0.39 & 0.43 & 0.41 & 0.51 & 0.46 & 0.53 & 0.47 & 0.54 & 0.50 \\
  & & MLP$^{2b}$    & \textbf{0.37} & 0.38 & 0.41 & 0.40 & 0.49 & 0.45 & 0.54 & 0.48 & 0.55 & 0.51 \\

    \cmidrule(lr){2-13}
  & \multirow{2}{*}{\textbf{mean}$^{3a}$} 
    & linear$^{1b}$ & 0.66 & 0.49 & 0.68 & 0.50 & 0.72 & 0.53 & 0.69 & 0.54 & 0.69 & 0.57 \\
  & & MLP$^{2b}$    & 0.61 & 0.50 & 0.74 & 0.51 & 0.77 & 0.54 & 0.74 & 0.55 & 0.73 & 0.58 \\
\midrule
\multirow{6}{*}{\texttt{ETTh2}} 
  & \multirow{2}{*}{\textbf{none}$^{1a}$} 
    & linear$^{1b}$ & \textbf{0.19} & \textbf{0.27} & \textbf{0.24} & \textbf{0.30} & \textbf{0.39} & \textbf{0.40} & \textbf{0.43} & \textbf{0.43} & \textbf{0.47} & \textbf{0.47} \\
  & & MLP$^{2b}$    & 0.20 & \textbf{0.27} & 0.25 & 0.31 & 0.40 & \textbf{0.40} & 0.46 & 0.44 & 0.51 & 0.48 \\

  \cmidrule(lr){2-13}
  & \multirow{2}{*}{\textbf{last time step}$^{1a}$} 
    & linear$^{1b}$ & \textbf{0.19} & 0.28 & 0.25 & 0.31 & 0.40 & \textbf{0.40} & 0.44 & \textbf{0.43} & 0.49 & 0.48 \\
  & & MLP$^{2b}$    & 0.20 & 0.28 & 0.26 & 0.32 & 0.40 & \textbf{0.40} & 0.45 & 0.44 & 0.49 & 0.48 \\

  \cmidrule(lr){2-13}
  & \multirow{2}{*}{\textbf{mean}$^{1a}$} 
    & linear$^{1b}$ & 0.25 & 0.32 & 0.29 & 0.35 & 0.44 & 0.43 & 0.44 & 0.45 & 0.50 & 0.49 \\
  & & MLP$^{2b}$    & 0.25 & 0.33 & 0.30 & 0.35 & 0.44 & 0.43 & 0.46 & 0.45 & 0.50 & 0.49 \\
\bottomrule
\end{tabular}
}
\caption{Ablation results comparing pooling strategies and heads across ETTh1 and ETTh2. \textbf{Bolded} values denote best within each dataset and horizon.}
\label{tab:combined_probe_ablations}
\end{table}

From \Cref{tab:combined_probe_ablations}, we observe that no pooling (1a) combined with either a linear or MLP probe yields the best results on both \texttt{ETTh1} and \texttt{ETTh2}. Interestingly, we observe that for \texttt{ETTh1} using just the last time step's embedding (1b) yields competitive scores with an average increase of $8.7\%$ (MSE) and $4.2\%$ (MAE) when compared to no pooling (1a). Comparatively, mean pooling (1c) has an increase of $60.5\%$ (MSE) and $24.4\%$ (MAE) Similarly, for \texttt{ETTh2}, we observe that using the last time step embeddings (1b) has only a $0.85\%$ (MSE) and $1.33\%$ (MAE) increase in error, when compared to mean pooling which has a $9.32\%$ (MSE) and $8.49\%$ (MAE) increase in error.

This observation is in line with \citep{vjepa}, which demonstrated that using attentive probing to pool embeddings was empirically superior for downstream task performance compared to directly mean pooling the embeddings, which can potentially result in lossy, diffuse representations which fail to capture finer granularities in the data.

\subsubsection{Weather}
\begin{table}[H]
\centering
\resizebox{\textwidth}{!}{%
\begin{tabular}{llcccccccccccc}
\toprule
\textbf{Dataset} & \textbf{H} 
  & \multicolumn{2}{c}{\textbf{Informer}} 
  & \multicolumn{2}{c}{\textbf{FEDformer}} 
  & \multicolumn{2}{c}{\textbf{UniTS-ST}\textsuperscript{\dag}} 
  & \multicolumn{2}{c}{\textbf{MOMENT$_\text{LP}$}} 
  & \multicolumn{2}{c}{\textbf{CHARM}} \\ 
\cmidrule(lr){3-4} \cmidrule(lr){5-6} \cmidrule(lr){7-8} \cmidrule(lr){9-10} \cmidrule(lr){11-12}
& 
& \textbf{MSE} & \textbf{MAE} 
& \textbf{MSE} & \textbf{MAE} 
& \textbf{MSE} & \textbf{MAE} 
& \textbf{MSE} & \textbf{MAE}
& \textbf{MSE} & \textbf{MAE} \\
\midrule
\multirow{4}{*}{\textbf{Weather}} 
  & 96  & 0.300 & 0.384 & 0.217 & 0.296 & \textbf{0.149} & \textbf{0.198} & \underline{0.154} & 0.209 & 0.158 & \underline{0.199} \\
  & 192 & 0.598 & 0.544 & 0.276 & 0.336 & \underline{0.200} & \textbf{0.243} & \textbf{0.197} & 0.248 & 0.207 & \underline{0.246} \\
  & 336 & 0.578 & 0.523 & 0.339 & 0.380 & \underline{0.257} & \textbf{0.286} & \textbf{0.246} & \underline{0.285} & 0.265 & 0.287 \\
  & 672 & 1.059 & 0.741 & 0.403 & 0.428 & \underline{0.334} & \textbf{0.338} & \textbf{0.315} & \underline{0.336} & 0.347 & 0.340 \\
\midrule
& Avg & 0.634 & 0.548 & 0.309 & 0.360 & \underline{0.235} & \textbf{0.266} & \textbf{0.228} & 0.270 & 0.244 & \underline{0.268} \\
\bottomrule
\end{tabular}
}
\caption{Long-horizon forecasting results on the Weather dataset across different horizons (input length = 96). Lower is better. \textbf{Bold} = best, \underline{Underline} = second best.}
\label{tab:weather_results}
\end{table}

\subsubsection{Exchange Rate}
\begin{table}[H]
\centering

\resizebox{\textwidth}{!}{\begin{tabular}{ll*{8}{cc}}
\toprule
\textbf{Dataset} & \textbf{H} & \multicolumn{2}{c}{\textbf{Autoformer}} & \multicolumn{2}{c}{\textbf{Informer}} & \multicolumn{2}{c}{\textbf{LogTrans}} & \multicolumn{2}{c}{\textbf{Reformer}} & \multicolumn{2}{c}{\textbf{LSTNet}} & \multicolumn{2}{c}{\textbf{LSTM}} & \multicolumn{2}{c}{\textbf{TCN}} & \multicolumn{2}{c}{\textbf{CHARM}} \\
\cmidrule(lr){3-4} \cmidrule(lr){5-6} \cmidrule(lr){7-8} \cmidrule(lr){9-10} \cmidrule(lr){11-12} \cmidrule(lr){13-14} \cmidrule(lr){15-16} \cmidrule(lr){17-18}
& & \textbf{MSE} & \textbf{MAE} & \textbf{MSE} & \textbf{MAE} & \textbf{MSE} & \textbf{MAE} & \textbf{MSE} & \textbf{MAE} & \textbf{MSE} & \textbf{MAE} & \textbf{MSE} & \textbf{MAE} & \textbf{MSE} & \textbf{MAE} & \textbf{MSE} & \textbf{MAE} \\
\midrule
\multirow{4}{*}{\textbf{Exchange Rate}}
& 96  & \underline{0.197} & \underline{0.323} & 0.847 & 0.752 & 0.968 & 0.812 & 1.065 & 0.829 & 1.551 & 1.058 & 1.453 & 1.049 & 3.004 & 1.432 & \textbf{0.084} & \textbf{0.203} \\
& 192 & \underline{0.300} & \underline{0.369} & 1.204 & 0.895 & 1.040 & 0.851 & 1.188 & 0.906 & 1.477 & 1.028 & 1.846 & 1.179 & 3.048 & 1.444 & \textbf{0.182} & \textbf{0.302} \\
& 336 & \underline{0.509} & \underline{0.524} & 1.672 & 1.036 & 1.659 & 1.081 & 1.357 & 0.976 & 1.507 & 1.031 & 2.136 & 1.231 & 3.113 & 1.459 & \textbf{0.353} & \textbf{0.429} \\
& 720 & \underline{1.447} & \underline{0.941} & 2.478 & 1.310 & 1.941 & 1.127 & 1.510 & 1.016 & 2.285 & 1.243 & 2.984 & 1.427 & 3.150 & 1.458 & \textbf{0.929} & \textbf{0.727} \\
\midrule
& Avg & \underline{0.613} & \underline{0.539} & 1.550 & 0.998 & 1.402 & 0.968 & 1.280 & 0.932 & 1.705 & 1.090 & 2.105 & 1.222 & 3.079 & 1.448 & \textbf{0.387} & \textbf{0.415} \\
\bottomrule
\end{tabular}
}
\caption{
\\
Long horizon forecasting results on the Exchange Rate dataset across different horizons. Input length = $96$. \\
Lower is better. \textbf{Bold} = best, \underline{underline} = second best.
}
\label{tab:exchange_rate_results_all_models}
\end{table}

\subsubsection{Influenza Like Illness (ILI)}

\begin{table}[H]
\centering
\resizebox{\textwidth}{!}{
    \begin{tabular}{ll*{9}{cc}}
    \toprule
    \textbf{Dataset} & \textbf{H}
      & \multicolumn{2}{c}{\textbf{MOMENT\textsubscript{LP}}}
      & \multicolumn{2}{c}{\textbf{DLinear}}
      & \multicolumn{2}{c}{\textbf{Stationary}}
      & \multicolumn{2}{c}{\textbf{N‑BEATS}}
      & \multicolumn{2}{c}{\textbf{Autoformer}}
      & \multicolumn{2}{c}{\textbf{Reformer}}
      & \multicolumn{2}{c}{\textbf{LSTM}}
      & \multicolumn{2}{c}{\textbf{TCN}}
      & \multicolumn{2}{c}{\textbf{CHARM}} \\
    \cmidrule(lr){3-4}  \cmidrule(lr){5-6}  \cmidrule(lr){7-8}
    \cmidrule(lr){9-10} \cmidrule(lr){11-12} \cmidrule(lr){13-14}
    \cmidrule(lr){15-16} \cmidrule(lr){17-18} \cmidrule(lr){19-20}
    \multirow{4}{*}{\textbf{ILI}} & 24
      & 2.728 & 1.114
      & \underline{2.215} & \underline{1.081}
      & 2.294 & 0.945
      & 4.539 & 1.528
      & 3.483 & 1.287
      & 4.400 & 1.382
      & 5.914 & 1.734
      & 6.624 & 1.830
      & \textbf{1.805} & \textbf{0.807} \\
    & 36
      & 2.669 & 1.092
      & \underline{1.963} & \underline{0.963}
      & 1.825 & 0.848
      & 4.628 & 1.534
      & 3.103 & 1.148
      & 4.783 & 1.448
      & 6.631 & 1.845
      & 6.858 & 1.879
      & \textbf{1.754} & \textbf{0.797} \\
    & 48
      & 2.728 & 1.098
      & \underline{2.130} & \underline{1.024}
      & 2.010 & 0.900
      & 4.957 & 1.585
      & 2.669 & 1.085
      & 4.832 & 1.465
      & 6.736 & 1.857
      & 6.968 & 1.892
      & \textbf{1.699} & \textbf{0.820} \\
    & 60
      & 2.883 & 1.126
      & \underline{2.368} & \underline{1.096}
      & 2.178 & 0.963
      & 5.429 & 1.661
      & 2.770 & 1.125
      & 4.882 & 1.483
      & 6.870 & 1.879
      & 7.127 & 1.918
      & \textbf{1.740} & \textbf{0.838} \\
    \midrule
      & Avg 
      & 2.752 & 1.108
      & \underline{2.169} & \underline{1.041}
      & 2.077 & 0.914
      & 4.888 & 1.577
      & 3.006 & 1.161
      & 4.724 & 1.445
      & 6.538 & 1.829
      & 6.894 & 1.880
      & \textbf{1.750} & \textbf{0.816} \\
    \bottomrule
    \end{tabular}
}
\caption{Long-horizon forecasting results on the ILI dataset across different horizons (input length = 96). Lower is better. \textbf{Bold} = best, \underline{underline} = second best.}
\label{tab:illness_results}
\end{table}

\section{Ablations}
\subsection{Effect of Channel Aware Layers}
\label{app:ablation}
To outline the efficacy of our customized text based attention layers, we conduct ablations to assess the overall impact on the embedding quality by analyzing how classification, forecasting, and SSL proxy metrics like LIDAR \footnote{LIDAR \citep{thilak2023lidarsensinglinearprobing} is a proxy metric which has been empirically shown to have high correlation to downstream tasks, and serves as an effective surrogate objective to track during training. The LIDAR score is computed by calculating the entropy of the eigenspectrum distribution of the LDA matrix, which represents the effective rank of the embedding space.} evolve over the course of training. Note that for these ablations we use linear probes without any additional hyperparameter tuning.

\begin{figure}[!h]
    \centering

    \begin{subfigure}[t]{0.32\textwidth}
        \centering
        \includegraphics[width=\linewidth]{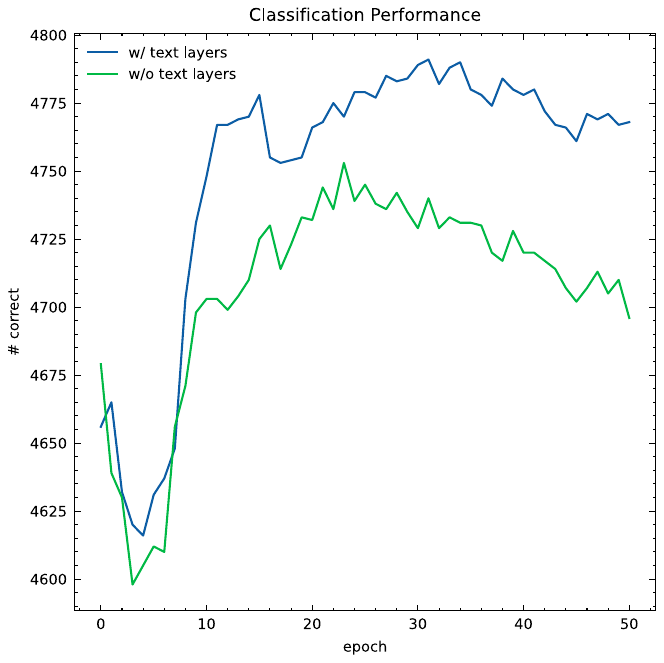}
        \caption{Classification Scores vs Epoch}
        \label{fig:text_ablation_uea}
    \end{subfigure}
    \hfill
    \begin{subfigure}[t]{0.32\textwidth}
        \centering
        \includegraphics[width=\linewidth]{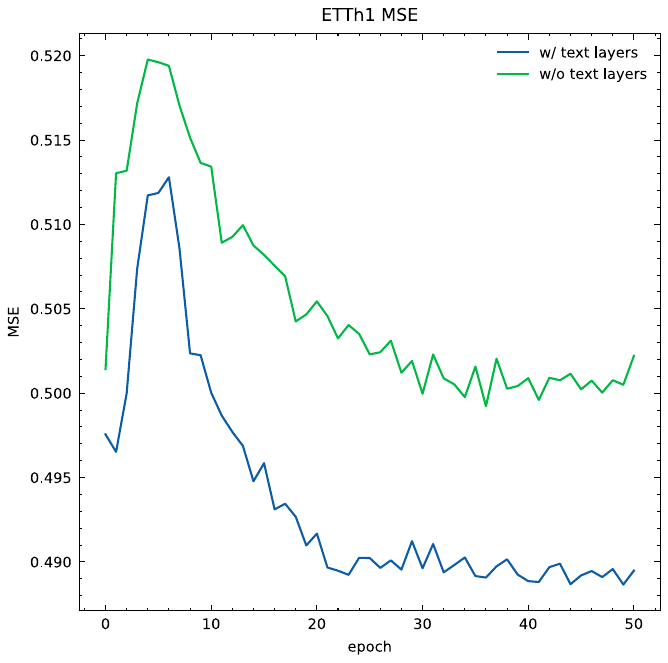}
        \caption{Forecasting Error vs Epoch}
        \label{fig:text_ablation_ett}
    \end{subfigure}
    \hfill
    \begin{subfigure}[t]{0.32\textwidth}
        \centering
        \includegraphics[width=\linewidth]{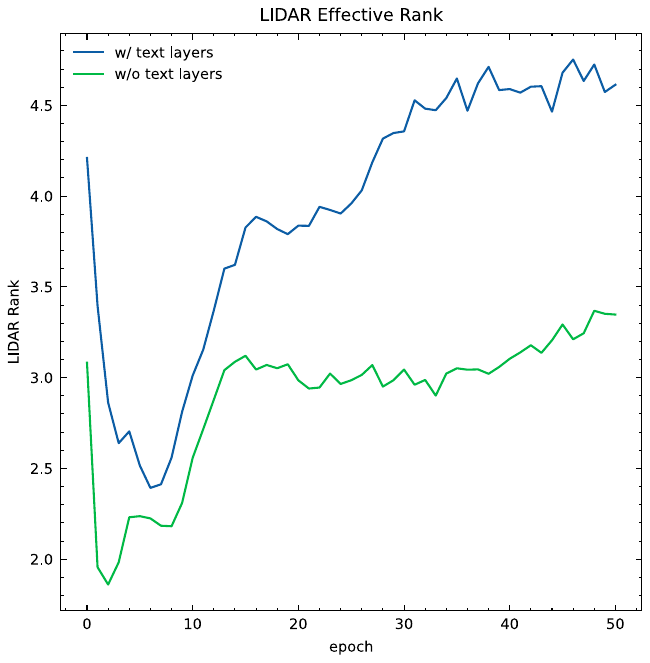}
        \caption{LIDAR Score vs Epoch}
        \label{fig:text_ablation_lidar}
    \end{subfigure}

    \caption{Ablation analysis of adding text layers across tasks: (a) classification (UEA), (b) forecasting (\texttt{ETTh1}), (c) representation learning (LIDAR).}
    \label{fig:text_ablation_all}
\end{figure}

\begin{table}[h]
\centering
\begin{tabular}{l>{\columncolor{lightblue}}ccc}
\toprule
\textbf{Metric} & \textbf{w/ text} & \textbf{w/o text} & \textbf{\% improvement} \\
\midrule
ETTh1 MSE ↓     & \textbf{0.48}  & 0.50  & +4.00\% \\
\# correct (UEA) ↑  & \textbf{4768}  & 4696  & +1.53\% \\
LIDAR ↑       & \textbf{4.61}  & 3.34  & +38.02\% \\
\bottomrule
\end{tabular}
\caption{Comparison of metrics between models trained with and without textual features. Positive \% indicates improvement.}
\end{table}

\section{Forecasting Visualizations}

\subsection{Illness}
\begin{figure}[H]
    \centering
    \includegraphics[width=\textwidth]{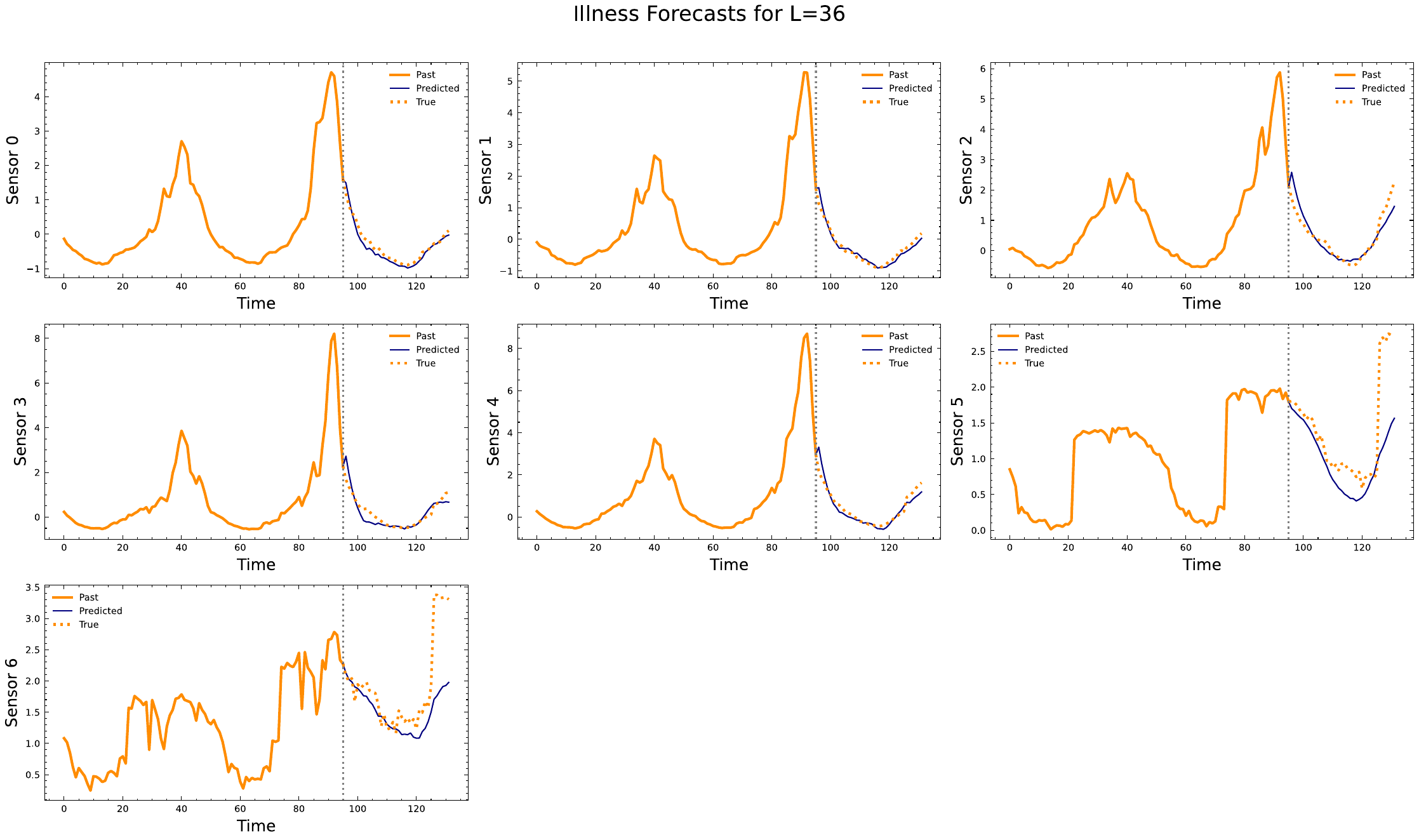}
    \caption{Illness Forecasts}
    \label{fig:illness-36}
\end{figure}

\begin{figure}[H]
    \centering
    \includegraphics[width=\textwidth]{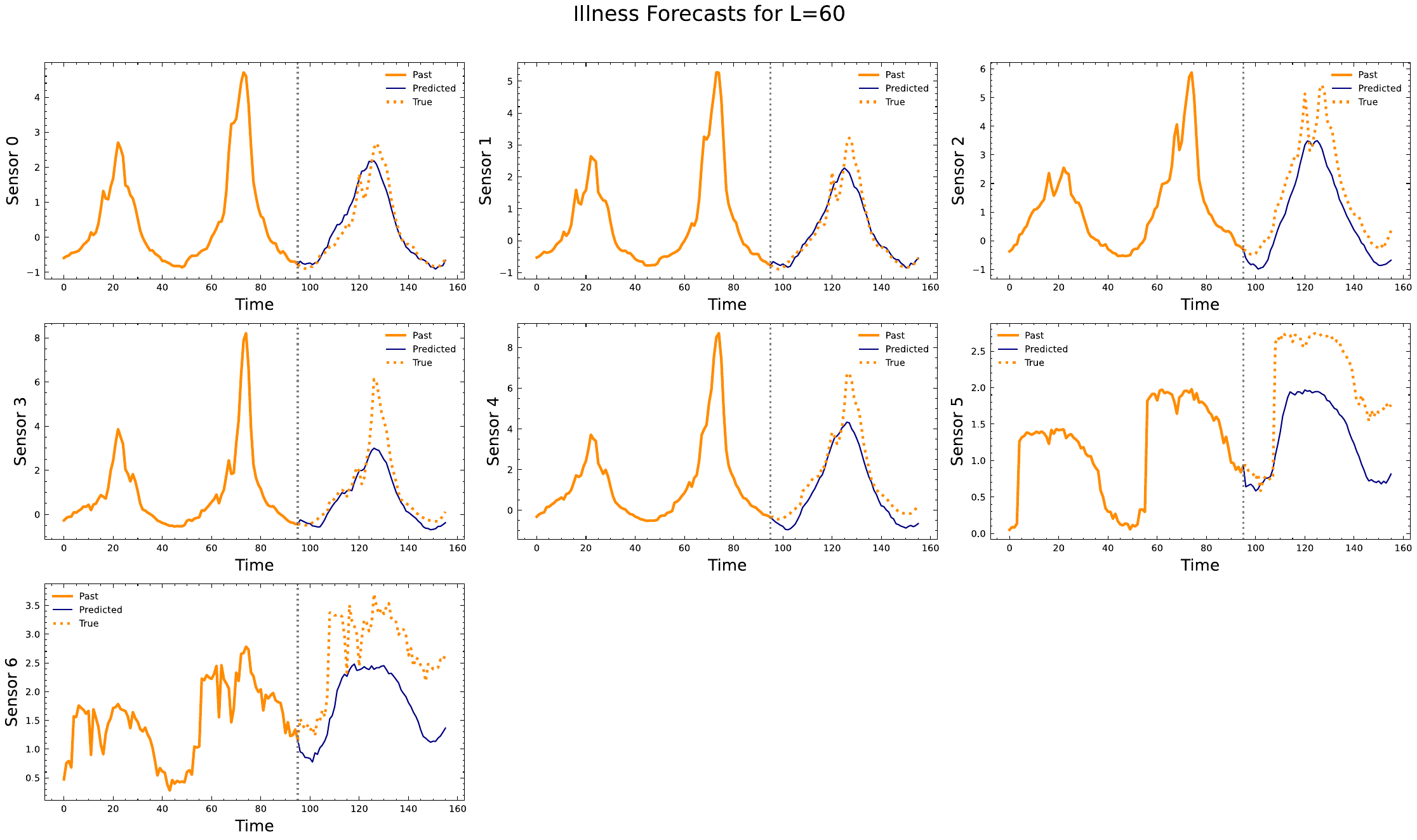}
    \caption{Illness Forecasts}
    \label{fig:illness-60}
\end{figure}

\subsection{ETTh}
\begin{figure}[H]
    \centering
    \includegraphics[width=\textwidth]{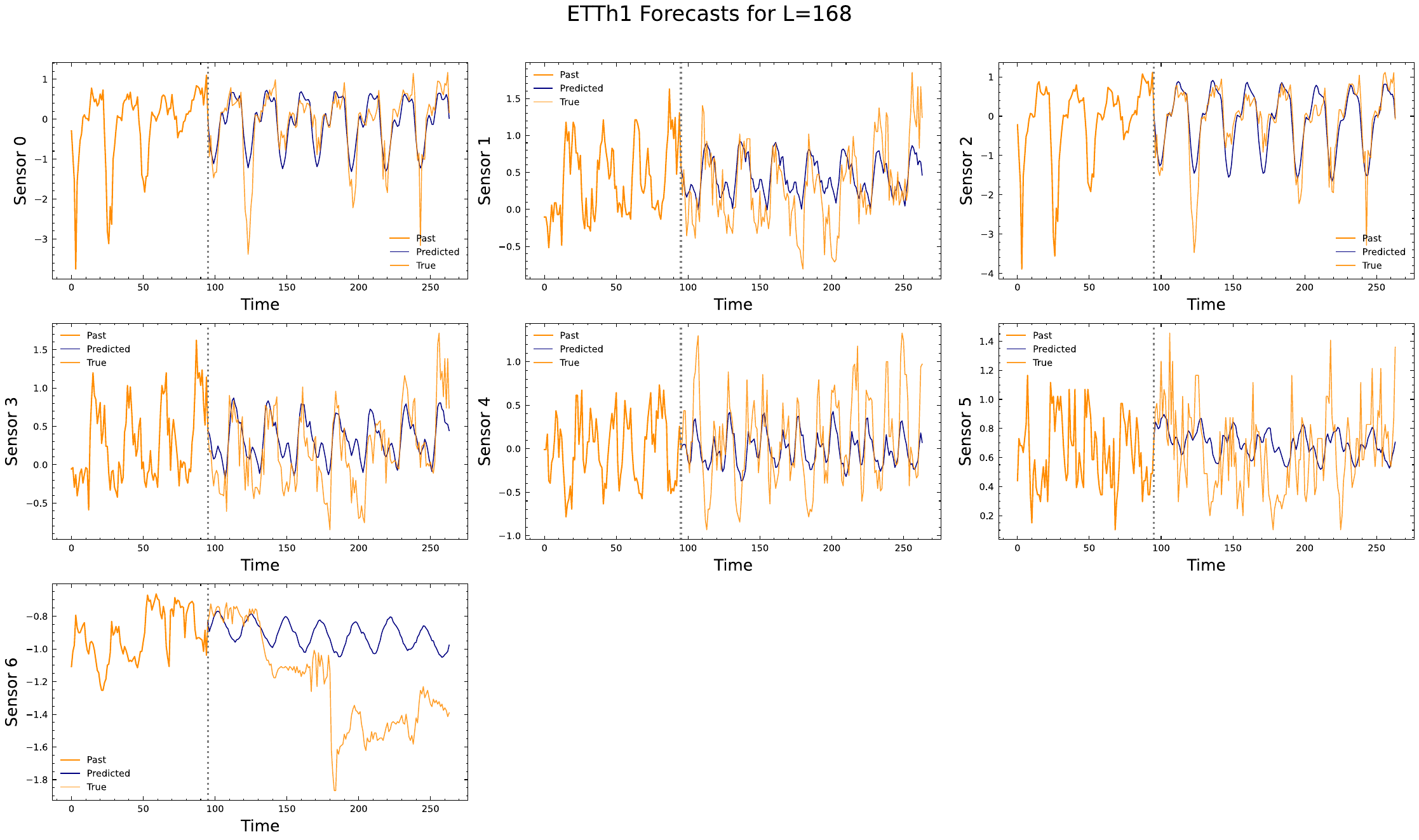}
    \caption{ETTh1 Forecasts}
    \label{fig:etth1-168}
\end{figure}

\begin{figure}[H]
    \centering
    \includegraphics[width=\textwidth]{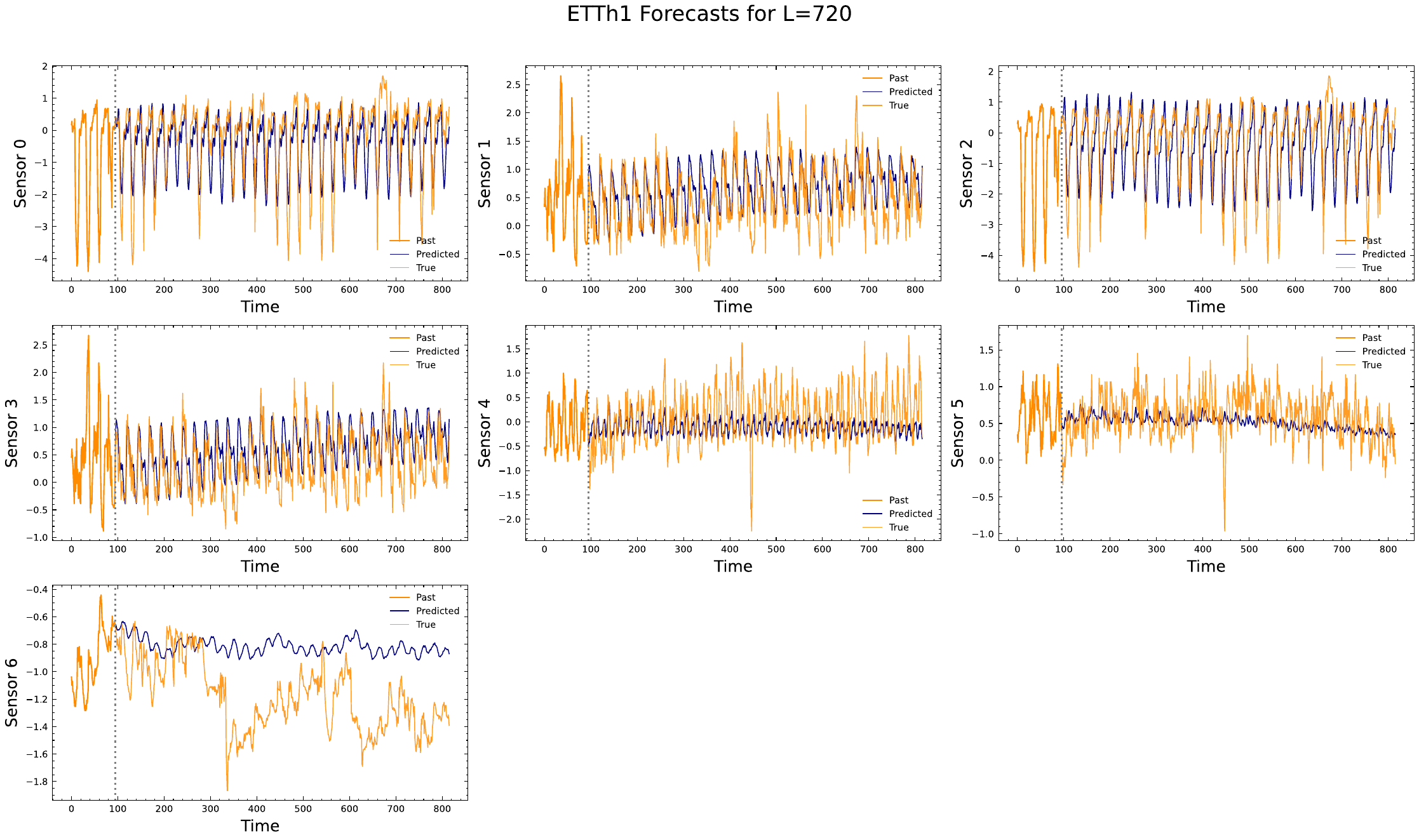}
    \caption{ETTh1 Forecasts}
    \label{fig:etth1-720}
\end{figure}

\begin{figure}[H]
    \centering
    \includegraphics[width=\textwidth]{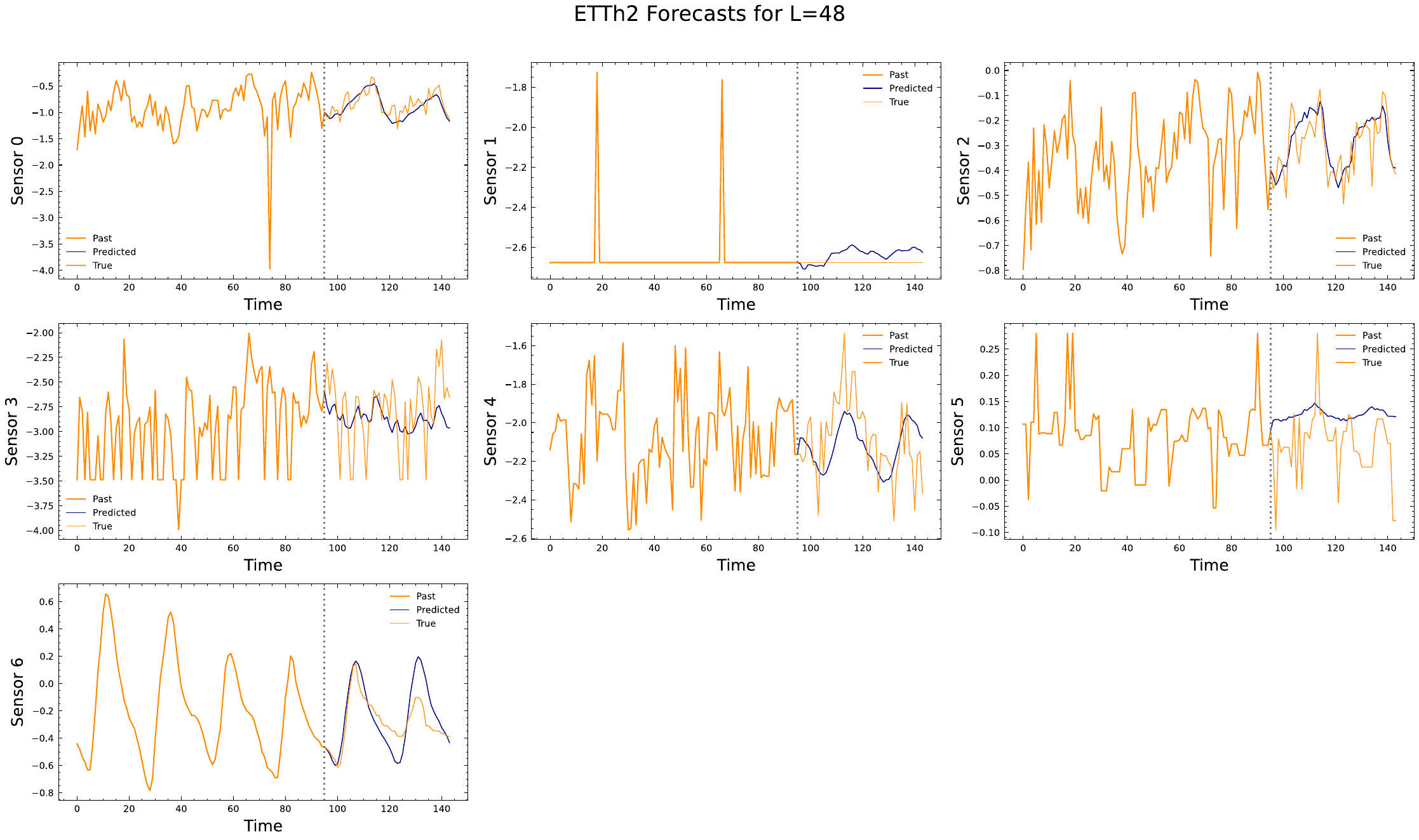}
    \caption{ETTh2 Forecasts}
    \label{fig:etth2-48}
\end{figure}

\begin{figure}[H]
    \centering
    \includegraphics[width=\textwidth]{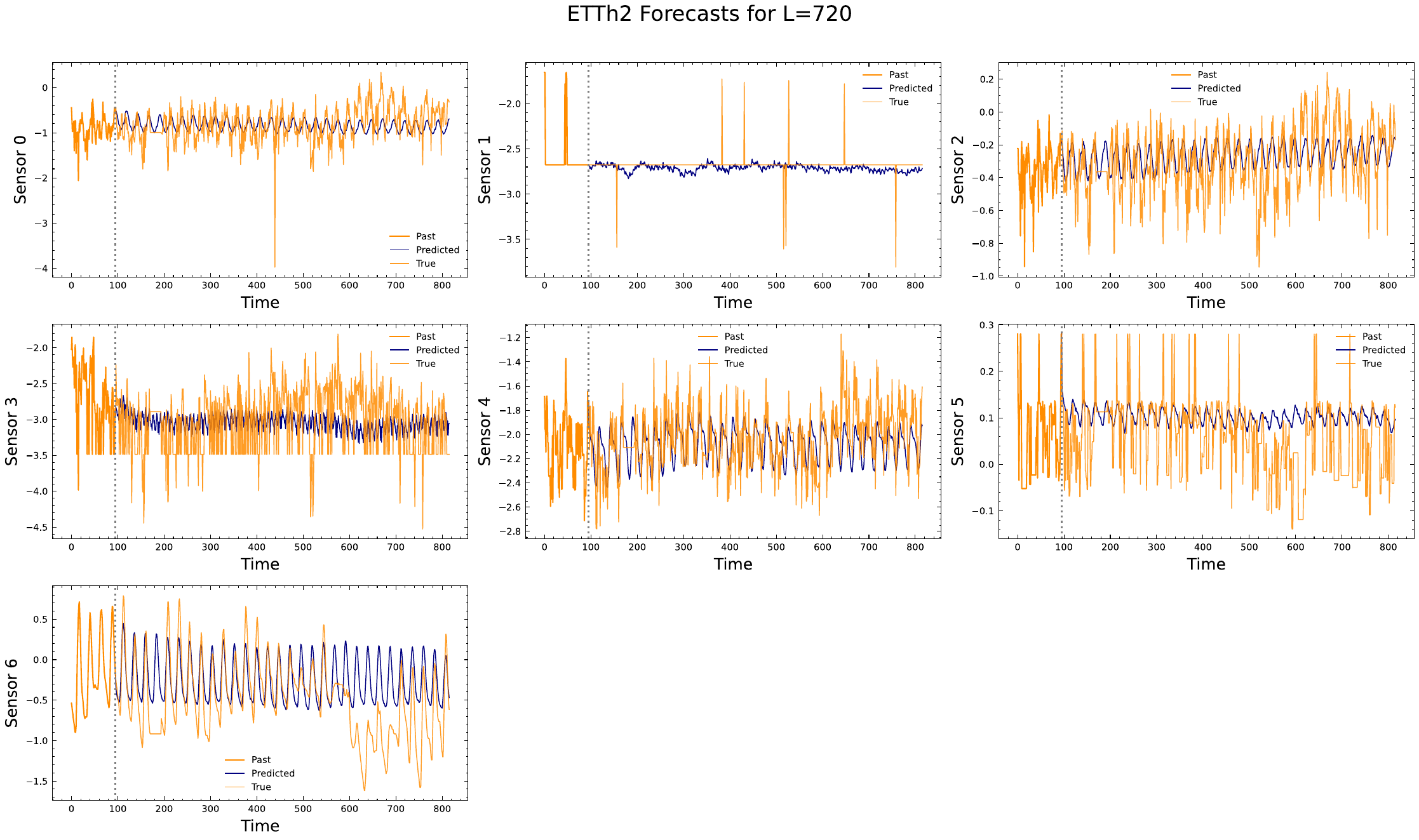}
    \caption{ETTh2 Forecasts}
    \label{fig:etth2-720}
\end{figure}

\subsection{ETTm}
\begin{figure}[H]
    \centering
    \includegraphics[width=\textwidth]{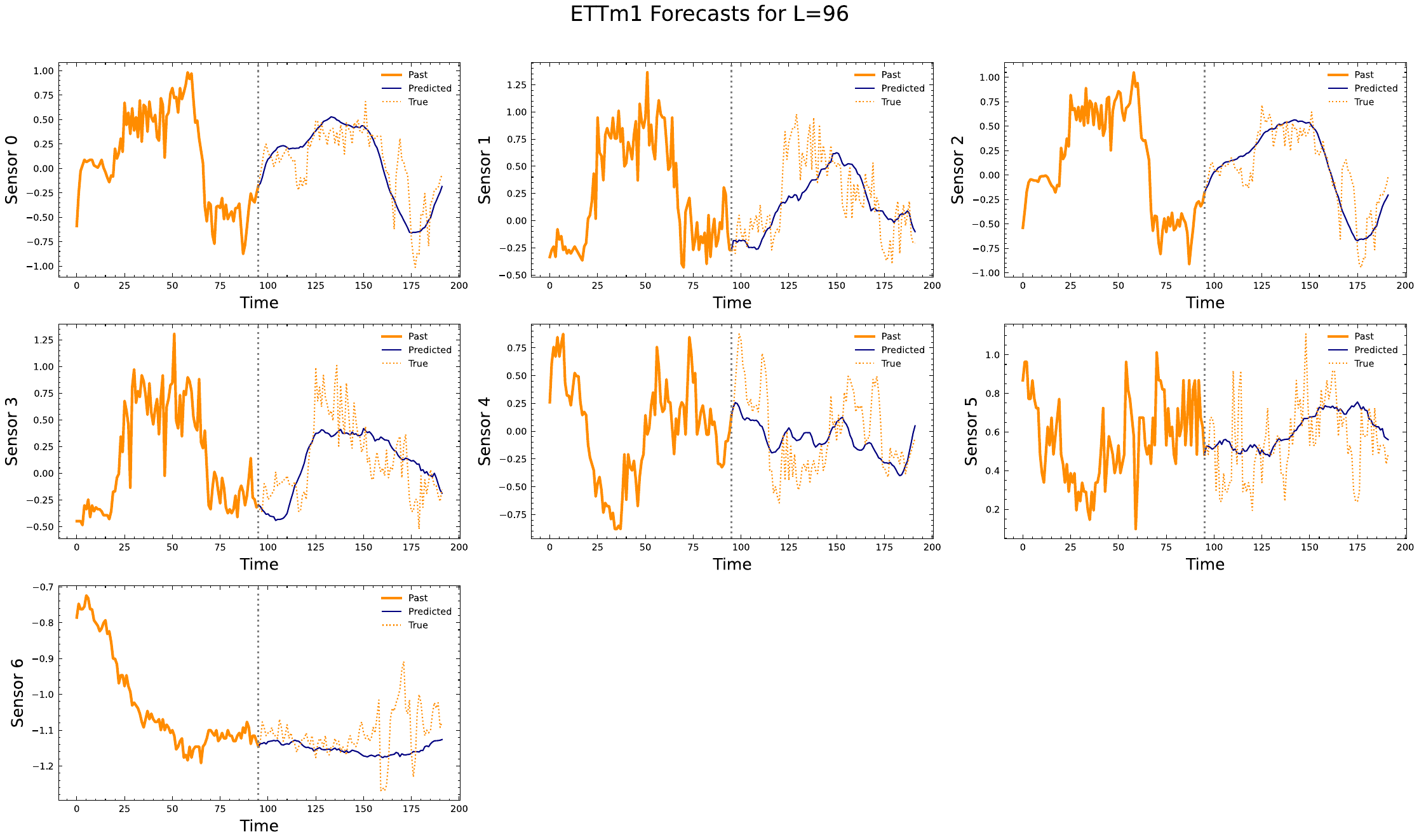}
    \caption{ETTm1 Forecasts}
    \label{fig:ettm1-96}
\end{figure}

\begin{figure}[H]
    \centering
    \includegraphics[width=\textwidth]{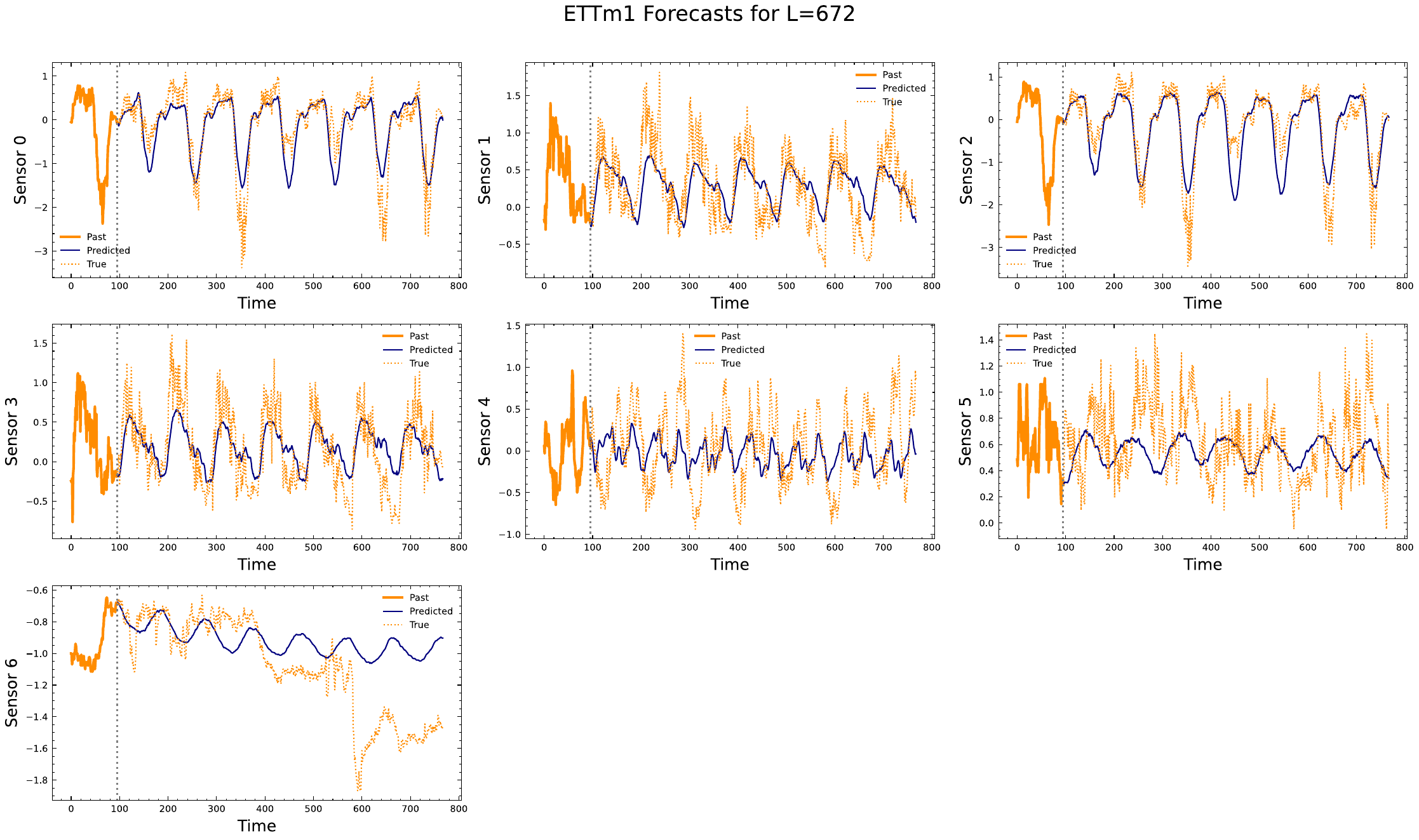}
    \caption{ETTm1 Forecasts}
    \label{fig:ettm1-672}
\end{figure}

% \begin{figure}[H]
%     \centering
%     \includegraphics[width=0.5\linewidth]{channel_gates.png}
%     \caption{Learned Channel Gating Matrix for a UEA Dataset}
%     \label{fig:enter-label}
% \end{figure}

% \begin{figure}[H]
%     \centering
%     \includegraphics[width=0.5\linewidth]{channel_gates_start.png}
%     \caption{Channel Gating Matrix At Start of Training}
%     \label{fig:enter-label}
% \end{figure}

% \begin{figure}
%     \centering
%     \includegraphics[width=0.5\linewidth]{tsne_start.png}
%     \caption{TSNE visualization of representations at start of training}
%     \label{fig:enter-label}
% \end{figure}

% \begin{figure}
%     \centering
%     \includegraphics[width=0.5\linewidth]{TSNE_end.png}
%     \caption{TSNE visualization of representations at end of training}
%     \label{fig:enter-label}
% \end{figure}

\clearpage

\section{Visualizations}
\subsection{Emergence of intra-class label separation} 
\label{app: heatmaps_label_sep}

To analyze how our model's embeddings evolve over training, we plot similarity heatmaps of our embeddings on labelled datasets.

We first obtain embeddings for a dataset by sampling a subset (approximately 50 samples) of the full dataset, while ensuring we have full label coverage. 
Given this embedding matrix $\mathbf{Z}\in \mathbb{R}^{N_t\times T\times C\times H}$, we obtain our mean-pooled embeddings $\bar{\mathbf{Z}}\in \mathbb{R}^{N_t\times H}$ by averaging over the channel and time dimension.

Finally, the $N_t\times N_t$ similarity matrix, $\mathbf{S}$ is obtained as follows:
\begin{align}
    \mathbf{S}_{i,j}=||\mathbf{Z}_{i,:}-\mathbf{Z}_{j,:}||_1
\end{align}

\begin{figure}[H]
    \centering

    \begin{subfigure}[t]{0.24\textwidth}
        \centering
        \includegraphics[width=\linewidth]{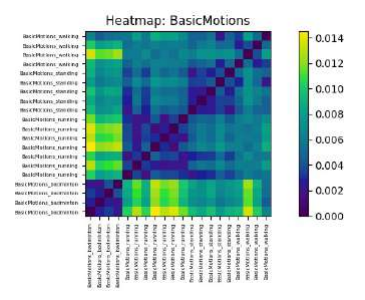}
        \caption{Epoch 0}
    \end{subfigure}
    \hfill
    \begin{subfigure}[t]{0.24\textwidth}
        \centering
        \includegraphics[width=\linewidth]{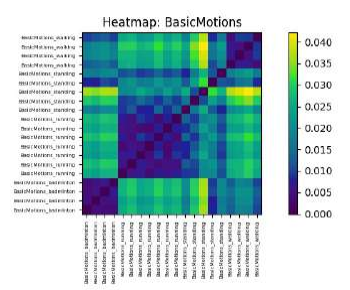}
        \caption{Epoch 3}
    \end{subfigure}
    \hfill
    \begin{subfigure}[t]{0.24\textwidth}
        \centering
        \includegraphics[width=\linewidth]{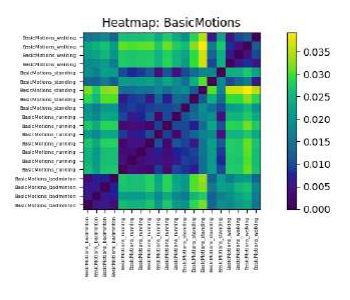}
        \caption{Epoch 6}
    \end{subfigure}
    \hfill
    \begin{subfigure}[t]{0.24\textwidth}
        \centering
        \includegraphics[width=\linewidth]{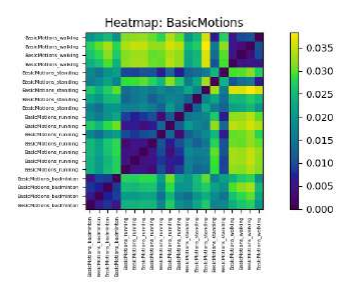}
        \caption{Epoch 9}
    \end{subfigure}
    \hfill
    \caption{Evolution of \texttt{BasicMotions} similarity heatmaps over training epochs}
    \label{app: bm_maps}
\end{figure}

\begin{figure}[!h]
    \centering

    \begin{subfigure}[t]{0.24\textwidth}
        \centering
        \includegraphics[width=\linewidth]{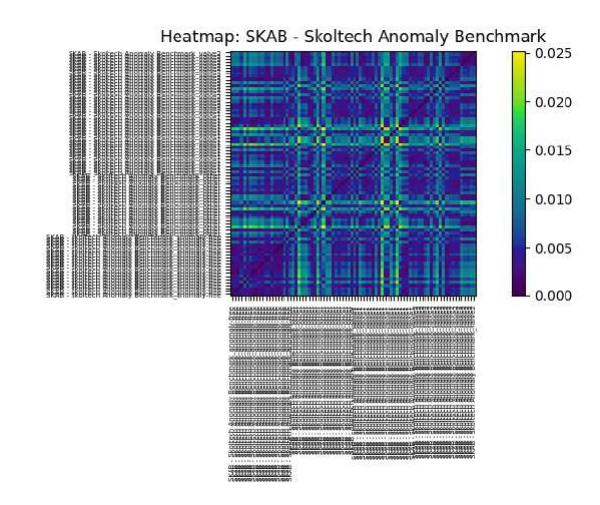}
        \caption{Epoch 0}
    \end{subfigure}
    \hfill
    \begin{subfigure}[t]{0.24\textwidth}
        \centering
        \includegraphics[width=\linewidth]{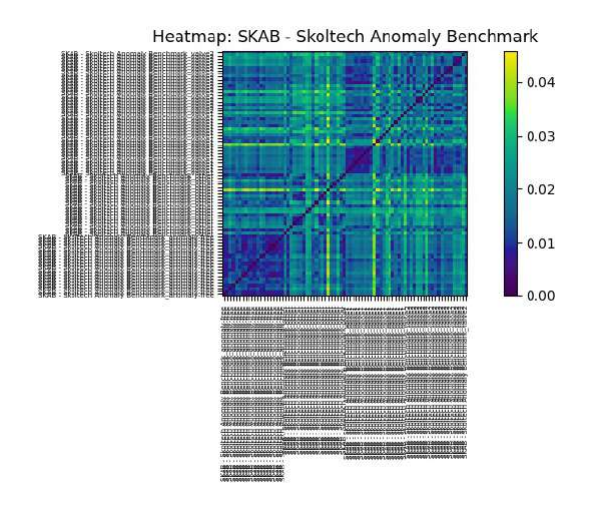}
        \caption{Epoch 3}
    \end{subfigure}
    \hfill
    \begin{subfigure}[t]{0.24\textwidth}
        \centering
        \includegraphics[width=\linewidth]{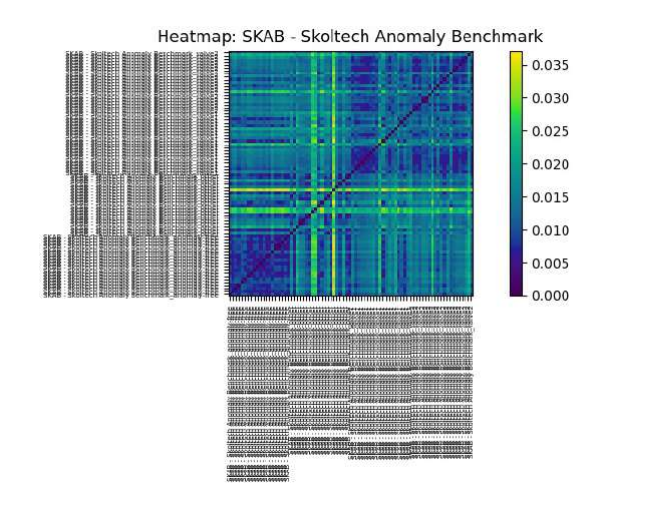}
        \caption{Epoch 6}
    \end{subfigure}
    \hfill
    \begin{subfigure}[t]{0.24\textwidth}
        \centering
        \includegraphics[width=\linewidth]{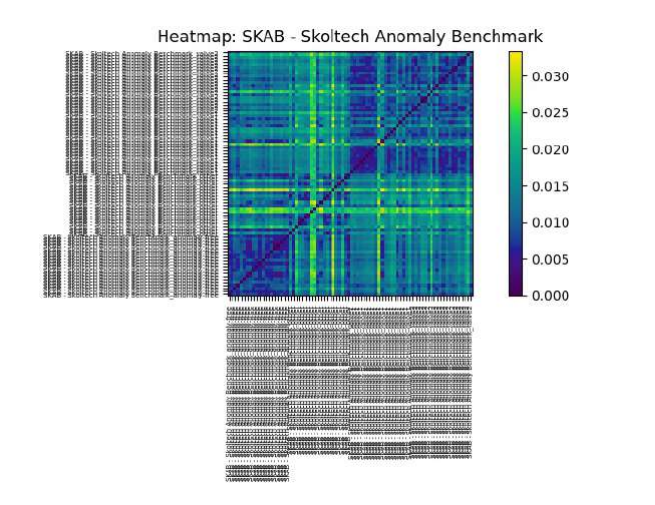}
        \caption{Epoch 9}
    \end{subfigure}
    \hfill
    \caption{Evolution of \texttt{Skoltech Anomaly Benchmark} similarity heatmaps over training epochs}
    \label{app: skab_maps}
    
\end{figure}

\begin{figure}[!h]
    \centering

    \begin{subfigure}[t]{0.24\textwidth}
        \centering
        \includegraphics[width=\linewidth]{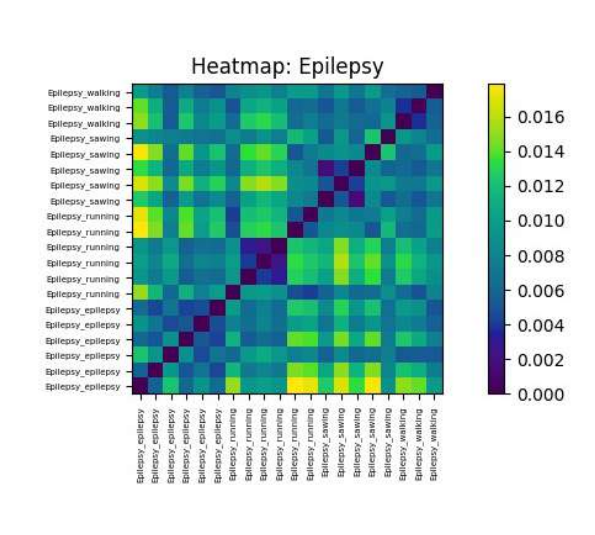}
        \caption{Epoch 0}
    \end{subfigure}
    \hfill
    \begin{subfigure}[t]{0.24\textwidth}
        \centering
        \includegraphics[width=\linewidth]{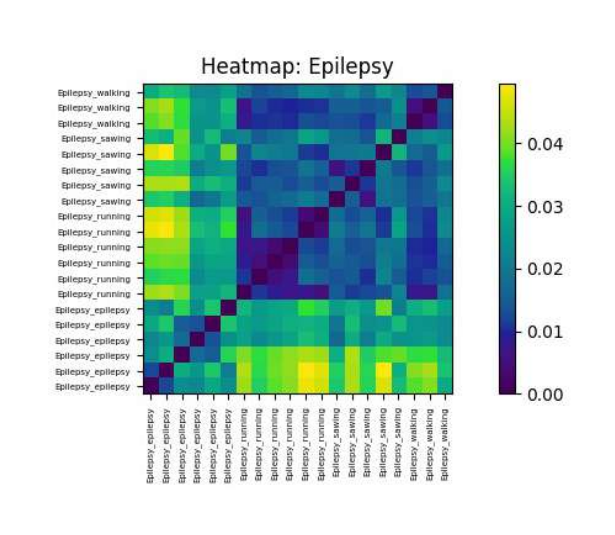}
        \caption{Epoch 3}
    \end{subfigure}
    \hfill
    \begin{subfigure}[t]{0.24\textwidth}
        \centering
        \includegraphics[width=\linewidth]{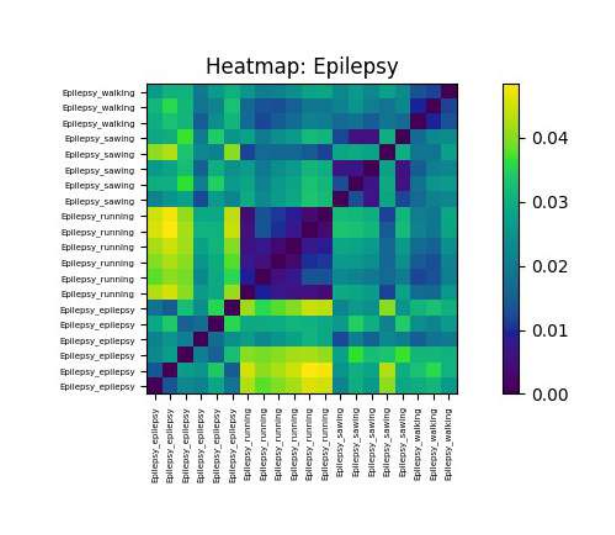}
        \caption{Epoch 6}
    \end{subfigure}
    \hfill
    \begin{subfigure}[t]{0.24\textwidth}
        \centering
        \includegraphics[width=\linewidth]{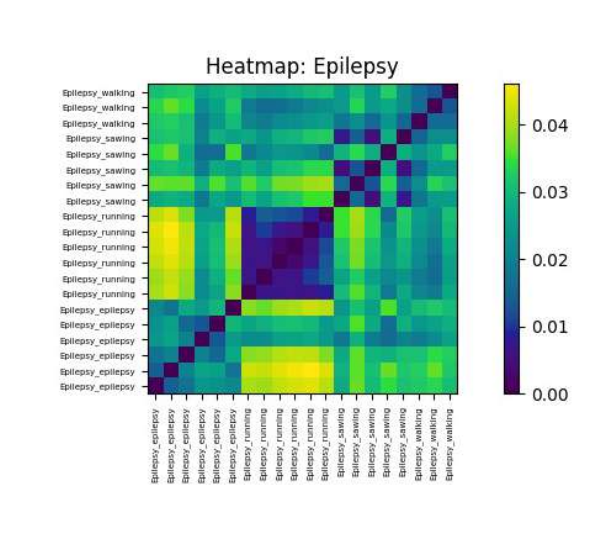}
        \caption{Epoch 9}
    \end{subfigure}
    \hfill
    \caption{Evolution of \texttt{Epilepsy} similarity heatmaps over training epochs}
    \label{app: ep_maps}
\end{figure}

We visualize the similarity matrix as a heatmap, as shown in \Cref{app: bm_maps,app: skab_maps,app: ep_maps}, and observe the emergence of structured clusters aligned with class labels. As training progresses, a block-diagonal structure\footnotemark becomes increasingly prominent, wherein samples sharing the same label exhibit reduced Euclidean separation compared to those from different classes. This pattern reflects a progressive tightening of intra-class representations, indicative of improved semantic organization in the learned embedding space.

\footnotetext{The heatmaps have a block structure because the labels are grouped together on each axis before plotting.}

\needspace{5\baselineskip}
\subsection{Evolution of Channel Gates}
\label{app:evolution_channel_gates}

In this section we aim to visualize how channel gates, as defined in Paragraph \Cref{channel_gates_desc}, evolve over the course of training our model. We plot the gating matrix, $\mathbf{G}_d$, for each dataset for different checkpoints. 

\begin{figure*}[!h]
    \centering
    \includegraphics[width=\linewidth, height=0.8\textheight, keepaspectratio]{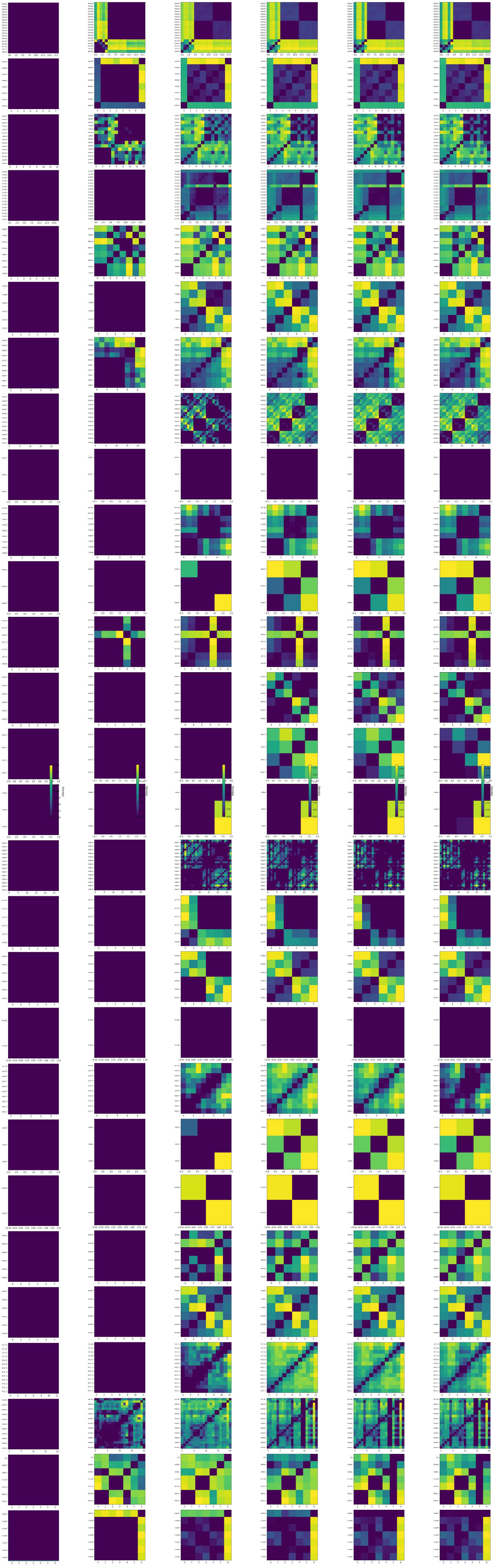}
    \caption{Evolution of inter channel gates during training. Checkpoints extracted at \texttt{epoch=0;step=49}, \texttt{epoch=0;step=499}, \texttt{epoch=0;step=999}, \texttt{epoch=2;step=49}, \texttt{epoch=6;step=49}, \texttt{epoch=8;step=49}. \\
    Each row represents a particular dataset. Each column represents a sampled checkpoint as training progresses. Each heatmap represents $\mathbf{G}_d$ for a particular dataset, which is a $C\times C$ matrix with values in $[0,1]$. Brighter colors on the heatmap represent \textbf{higher} gating values, i.e. decreased cross-channel interactions.
    }
    \label{fig:gates_evolve}
\end{figure*}

As illustrated in \Cref{fig:gates_evolve}, the inter-channel gating mechanism enables the model to dynamically modulate attention across channels, selectively emphasizing or suppressing information based on configurations that minimize the self-supervised learning (SSL) loss. We also empirically observe that the regularization loss begins to increase after an initial decline which suggests that after a certain point the model's embeddings require richer contextual information to continue improving.

\begin{figure*}[!h]
    \centering
    \includegraphics[width=\linewidth]{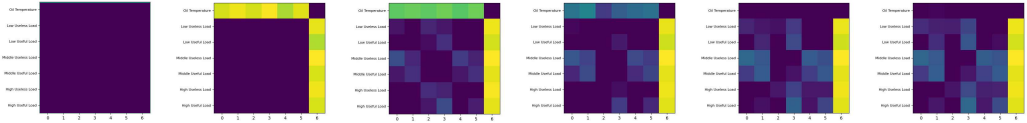}
    \caption{Evolution of Channel Gates for the \texttt{ETT} Dataset}
    \label{fig:ett_evolve_2}
\end{figure*}

The \texttt{ETT} dataset introduced by \citep{zho:21} comprises seven variables: \texttt{High Useful Load}, \texttt{Middle Useful Load}, \texttt{Low Useful Load}, \texttt{High Useless Load}, \texttt{Middle Useless Load}, \texttt{Low Useless Load}, and \texttt{Oil Temperature}. Among these, \texttt{Oil Temperature} serves as the target variable, with the remaining six acting as input features. During training, we observe a notable evolution in the learned channel gating patterns. Initially, the \texttt{Oil Temperature} channel does not attend to any other inputs, as indicated by its high gating values across all dimensions in \Cref{fig:ett_evolve_2}. However, as training progresses, this channel begins to incorporate information from all other variables. Interestingly, this behavior is asymmetric: while the target channel attends to all input features, the reverse does not occur—the other channels do not attend to \texttt{Oil Temperature}. This asymmetry manifests as a distinctive row-column pattern in the gating matrix and aligns with the underlying data semantics, where the target variable is causally influenced by the independent variables but not vice versa. These observations suggest that introducing learnable gating mechanisms can reveal interpretable, directional dependencies between variables which also increases model interpretability.

\end{document}